\documentclass[unnumsec,webpdf,contemporary,large, noend]{oup-authoring-template}
\graphicspath{{}}

\theoremstyle{thmstyleone}%
\theoremstyle{thmstyletwo}%
\theoremstyle{thmstylethree}%

\usepackage{hyperref}
\hypersetup{
    colorlinks=true,
    linkcolor=cyan,  
    urlcolor=cyan,   
    citecolor=cyan   
}
\usepackage{algorithmicx}
\usepackage{algorithm}
\usepackage{caption}
\usepackage{babel}
\usepackage{hyperref}
\usepackage{graphicx}
\usepackage{subcaption}
\usepackage{booktabs}
\usepackage{makecell}
\usepackage{graphicx}
\usepackage{setspace}
\usepackage{tcolorbox}

\begin{document}

\journaltitle{Briefings in Bioinformatics}
\DOI{DOI HERE}
\copyrightyear{2024}
\pubyear{2024}
\access{Advance Access Publication Date: Day Month Year}
\appnotes{Problem Solving Protocol}

\firstpage{1}


\title[DIG-Mol: GNN-based predictor for Molecular Property]{Contrastive Dual-Interaction Graph Neural Network for Molecular Property Prediction}

\author[1, $\dagger$]{Zexing Zhao}
\author[2,3 $\dagger$ ]{Guangsi Shi}
\author[1]{Xiaopeng Wu}
\author[4]{Ruohua Ren}
\author[1,$\ast$]{Xiaojun Gao}
\author[5,$\ast$]{Fuyi Li}

\authormark{Zhao et al.}

\address[1]{\orgdiv{College of Mechanical and Electronic Engineering}, \orgname{Northwest A\&F University}, \orgaddress{\street{Yangling}, \postcode{712100}, \country{China}}}
\address[2]{\orgdiv{Department of Chemical and Biological Engineering}, \orgname{Monash University}, \orgaddress{\street{Clayton}, \postcode{38000}, \state{VIC}, \country{Australia}}}
\address[3]{\orgdiv{Bosch Corporate Research}, \orgname{Bosch}, \orgaddress{\street{Changning}, \postcode{200335}, \state{Shanghai}, \country{China}}}
\address[4]{\orgdiv{Department of Materials Science and Engineering}, \orgname{Monash University}, \orgaddress{\street{Clayton}, \postcode{38000}, \state{VIC}, \country{Australia}}}
\address[5]{\orgdiv{College of Information Engineering}, \orgname{Northwest A\&F University}, \orgaddress{\street{Yangling}, \postcode{712100}, \country{China}}}

\corresp[$\dagger$]{The first two authors contribute equally
\\}

\corresp[$\ast$]{Corresponding author.
Xiaojun Gao: \href{email:xiaojungao@nwafu.edu.cn}{xiaojungao@nwafu.edu.cn} and
Fuyi Li: \href{email:fuyi.li@nwafu.edu.cn}{fuyi.li@nwafu.edu.cn}}

\received{Date}{0}{Year}
\revised{Date}{0}{Year}
\accepted{Date}{0}{Year}


\abstract{
Molecular property prediction is a pivotal component in the realm of AI-driven drug discovery and molecular representation learning. Despite recent progress, existing methods struggle with challenges such as limited generalization capabilities and inadequate representation learning from unlabeled data, particularly in tasks specific to molecular structures. To address these limitations, we introduce DIG-Mol, a novel self-supervised graph neural network framework for molecular property prediction. This architecture harnesses the power of contrastive learning, employing a dual-interaction mechanism and a unique molecular graph augmentation strategy. DIG-Mol integrates a momentum distillation network with two interlinked networks to refine molecular representations effectively. The framework’s ability to distill critical information about molecular structures and higher-order semantics is bolstered by minimizing contrastive loss. Our extensive experimental evaluations across various molecular property prediction tasks establish DIG-Mol’s state-of-the-art performance. In addition to demonstrating exceptional transferability in few-shot learning scenarios, our visualizations highlight DIG-Mol’s enhanced interpretability and representation capabilities. These findings confirm the effectiveness of our approach in overcoming the challenges faced by conventional methods, marking a significant advancement in molecular property prediction.
\\
\textbf{Availability: }The trained models and source code of DIG-Mol are publicly available at \url{http://github.com/ZeXingZ/DIG-Mol}.
}



\maketitle

\section{Introduction}

Navigating the challenging landscape of drug discovery and design, where target uncertainties and high attrition rates lead to extensive costs and timeframes, is a pressing concern \cite{intro1, intro2}. The journey from initial discovery to market launch can exceed a decade and cost upwards of 1 billion USD \cite{intro3, intro4}. Within this context, molecular representation learning has emerged as a pivotal tool, revolutionizing the understanding of molecular structures and properties. Central to this endeavor is molecular property prediction, serving as a critical benchmark for the effectiveness of representation learning techniques. It plays an integral role in key processes such as target identification \cite{intro6, intro7}, designing small molecules and new biological entities \cite{intro9}, and drug repurposing \cite{intro12}. Conventional wet-lab methods fall short of delivering accurate molecular representations for a vast number of molecules due to their impractical nature \cite{intro15}. Computational approaches have gained prominence as a remedy, accelerating drug discovery by efficiently acquiring molecular representations and bolstering AI-assisted drug discovery efforts \cite{fingerprints1, intro18}.

The advancement of computational methods in drug development has enabled the accumulation of extensive datasets. These datasets are integral for constructing quantitative structure-activity/property relationship models (QSAR/QSPR) and advancing molecular feature engineering \cite{intro18, lfy4}. Conventional methods primarily relied on manually crafted molecular features. However, the emergence of AI-driven, data-centric approaches in molecular representation learning has significantly transformed the landscape, addressing the constraints inherent in domain knowledge-based methodologies \cite{nlp1, intro26}. Among these developments, graph-based representations, particularly graph neural network (GNN) models, have become increasingly prominent. Their effectiveness lies in their ability to comprehensively capture molecular topology, rendering them highly effective for molecular analysis \cite{intro25, intro35}. By employing molecular graphs as input, GNN models have consistently delivered favorable outcomes in various molecular studies \cite{rev1, gcn2}.

Despite the notable achievements of graph neural networks in molecular property prediction, the limited availability of labeled data remains a significant challenge \cite{intro16}. To address this, recent research has gravitated towards self-supervised methods, which capitalize on extensive unlabeled molecular datasets for pre-training models. These methods have shown promising results in achieving robust performance, even when labeled data is scarce. A prominent approach in this realm is Graph Contrastive Learning (GCL), which utilizes data augmentation to create comparable sample pairs for contrastive learning \cite{gcl1}. This approach has been increasingly refined, with ongoing efforts to enhance data augmentation strategies and improve GNN encoders. Such enhancements often include integrating attention mechanisms and domain-specific knowledge into the learning process, aiming to boost the overall efficacy and accuracy of the models \cite{kg1, gat2}.

In spite of the significant strides in molecular representation learning and property prediction achieved by Graph Contrastive Learning (GCL) methods, several challenges remain unresolved. A primary issue with conventional GCL approaches is their reliance on contrastive learning between positive and negative sample pairs derived from the same molecules. This approach, though foundational, often falls short in efficiency and breadth of learning, as it focuses narrowly on the similarity between two augmented molecular graphs. Such a strategy leads to a rigid learning mechanism that may not fully capture the complexities of diverse and intricate molecular datasets. Furthermore, the effectiveness of contrastive representation learning heavily depends on the quality of the augmented contrastive pairs \cite{gcl1, cas}. Existing methods typically generate these pairs through either the graph attention mechanism or random masking of nodes and edges \cite{gat2,nat1}. While attention-guided masking can identify key substructures, it may overlook the individual importance of atoms. Conversely, random masking is based on chance, potentially missing critical molecular motifs. These current practices tend to lose molecular specificity during the augmentation process. Overlooking key interatomic interactions could have unpredictable impacts since some important atoms are removed. Therefore, there is a need for more nuanced augmentation strategies that maintain molecular specificity and better reflect the true nature of molecular structures and properties.

Throughout the augmentation operations, simplistic node masking and bond deletion may prove too rudimentary for scenarios where a solitary atom or chemical bond significantly influences molecular properties. Bond deletion, for instance, can entirely nullify the interaction between two atoms, potentially resulting in the isolation of pivotal atoms and the omission of critical structural information within the molecular graph \cite{aug1}.

The current augmentation techniques used in molecular graph contrastive learning, such as simplistic node masking and bond deletion, can be overly rudimentary for scenarios where individual atoms or bonds play a significant role in determining molecular properties. For example, deleting a bond could disrupt the interaction between two atoms entirely, leading to the loss of essential atoms and the exclusion of vital structural information from the molecular graph \cite{aug1}. These rudimentary augmentation methods can hinder the model’s ability to fully grasp the nuances of molecular data and limit its capability to generate accurate molecular representations. Consequently, the model or framework may become overly generic, losing its molecular specificity and largely depending on basic graphical and geometric properties, thereby failing to capture the intricate details that are crucial for understanding molecular structures and functions.

We introduce DIG-Mol, a \begin{math}
    \mathbf{D}\end{math}ual-\begin{math}
       \mathbf{I} 
    \end{math}nteraction \begin{math}
        \mathbf{G}
    \end{math}raph contrastive neural network for \begin{math}
        \mathbf{Mol}
    \end{math}ecular property prediction by fusing a graph diffusion network with a momentum distillation pseudo-siamese architecture for advanced graph augmentation. DIG-Mol optimizes the contrasting mechanism with dual-interaction contrastiveness, graph-interaction and encoder interaction, advancing the comprehensive understanding of molecular representations for property prediction. This is achieved by generating robust molecular representations from the identical molecular graph, traversing through various combinations of augmented graphs and networks. The momentum distillation pseudo-siamese network enhances the model's learning efficacy by implementing the "supervisor-labor" paradigm, where the online network is guided by a meticulously supervising target network that distills essential features from historical observations. Innovatively, DIG-Mol employs a molecule-specific graph augmentation strategy that preserves the directional message passing between atoms, avoiding the limitations of traditional attention-guided or random masking. This allows for the capture of both local and global potential information and the identification of significant molecular substructures, improving the accuracy of molecular property prediction. The strategic combination of node masking and unidirectional bond deletion, alongside the graph diffusion process, enables DIG-Mol to capture vital molecule-specific information. 

Extensive benchmarking against contemporary baseline models showcases DIG-Mol's superior performance and generalization capabilities in molecular property prediction tasks. By leveraging self-supervised learning and intelligently engineered network interactions, DIG-Mol sets a new standard in the domain, offering interpretable, high-quality molecular representations that significantly outperform competitive baselines. The key contributions of DIG-Mol lie in its innovative architecture, self-supervised learning strategy, and the capacity to provide valuable interpretability for molecular analysis.

\section{Materials and methodology}

\begin{figure*}[htb]
	\centering
	\includegraphics[scale=0.099]{Fig.1.pdf}
	\caption{ The overarching architecture of DIG-Mol, a two-stage contrastive learning framework. Figure 1. depicts the primary components of the framework, encompassing five key submodules: A. Overview of DIG-Mol pre-training process. B. Overview of DIG-Mol finetuning stage. C. Potential applications of molecular property prediction. D. Computational logic of various contrastive losses in DIG-Mol. E. Legends of data streams and other contents in the DIG-Mol architecture.}
	\label{fig:Fig 1.}
\end{figure*}

\subsection{Problem definition}

The goal of molecular property prediction, when provided with a set of molecule SMILES sequences, is to translate these sequences into molecular graphs \begin{math}
    \mathcal{D}_{\mathcal{G}}=\left \{ \mathcal{G}_{1},\mathcal{G}_{2},...,\mathcal{G}_{n}  \right \} 
\end{math} and employ a graph neural network (GNN) framework \begin{math}
    f\left ( \mathcal{G}_{i}  \right ) 
\end{math} to learn informative molecular representations for predicting corresponding labels \begin{math}
    \mathcal{Y}=\left \{ \mathcal{Y}_{1},\mathcal{Y}_{2},...,\mathcal{Y}_{n}  \right \} 
\end{math}. This process is represented as:
\begin{align}
    \label{eq:Eq1}
    \mathcal{D}=\left \{ \mathcal{S}_{1},\mathcal{S}_{2}\dots \mathcal{S}_{n} \right \}\overset{translate}{\Longrightarrow} \mathcal{D}_{\mathcal{G}}=\left \{ \mathcal{G}_{1},\mathcal{G}_{2}\dots \mathcal{G}_{n} \right \}
\end{align}
\begin{align}
    \label{eq:Eq2}
     Predict: \mathcal{Y}_{i} =f\left ( \mathcal{G}_{i}  \right ) ,\;\mathcal{G}_{i} \in \left \{ \mathcal{G}_{1},\mathcal{G}_{2},...,\mathcal{G}_{n}  \right \}
\end{align}
GNNs specialize in processing graph-structured data to generate node representations, with an encoder \begin{math}g\end{math} taking a graph \begin{math}\mathcal{G = (X,A)}\end{math} as input — where \begin{math}\mathcal X\end{math} denotes node attributes and \begin{math}\mathcal A\end{math} means the adjacent matrix.

For clarity, we provided the detailed full GNN iteration process and notations used in this study in Appendix \textup{I}, Appendix \textup{II}, and Supplementary Table \ref{tab:Atab 1.}.

\subsection{Pretraining datasets}
During the pre-training phase, DIG-Mol utilized a substantial dataset comprising approximately 10 million unique, unlabeled molecular SMILES sourced from the PubChem database \cite{pubchem}. We applied the ChemBERT framework and utilized RDKit for the conversion of these SMILES strings into molecular graphs, from which we extracted relevant chemical features. The dataset was then randomly split, allocating 95\% for training and 5\% for validation purposes.

\subsection{The overall framework of DIG-MOL}

DIG-MOL is a self-supervised contrastive graph deep learning framework tailored for learning molecular graph representations and predicting molecular properties. DIG-Mol employs a novel contrastive learning strategy, utilizing a dual-pathway network comprising both target and online networks, each with a graph neural network (GNN) encoder and a projection head.

Initially, molecular SMILES strings are converted into molecular graphs, followed by the generation of augmented graph pairs. These pairs are fed into both network pathways, creating four molecular representations from the augmented graphs and networks, which are used in the computation of contrastive loss. After pretraining, the GNN encoder is adapted to the fine-tuning stage for downstream tasks in drug discovery. The workflow from molecular data transformation to potential application is depicted in Figure 1, with further architecture details provided in Appendix \textup{III}, Algorithm \ref{alg:Alg 1.}.

\subsubsection{A: Graph augmentation strategy}

In this study, we employed a novel graph augmentation strategy that generated high-quality positive pairs from corrupt molecular graphs. This strategy expands a \begin{math}
\mathcal{G}_{i}    
\end{math} into two augmented versions, \begin{math}\mathcal{G}_{i}^{1}    
\end{math} and \begin{math}\mathcal{G}_{i}^{2}    
\end{math}, which retain correlated structures yet differ in specific areas. These are regarded as positive pairs for contrastive learning, while augmented graphs from different originals in the same mini-batch serve as negative pairs.

\begin{figure*}[htb]
	\centering
	\includegraphics[scale=0.6]{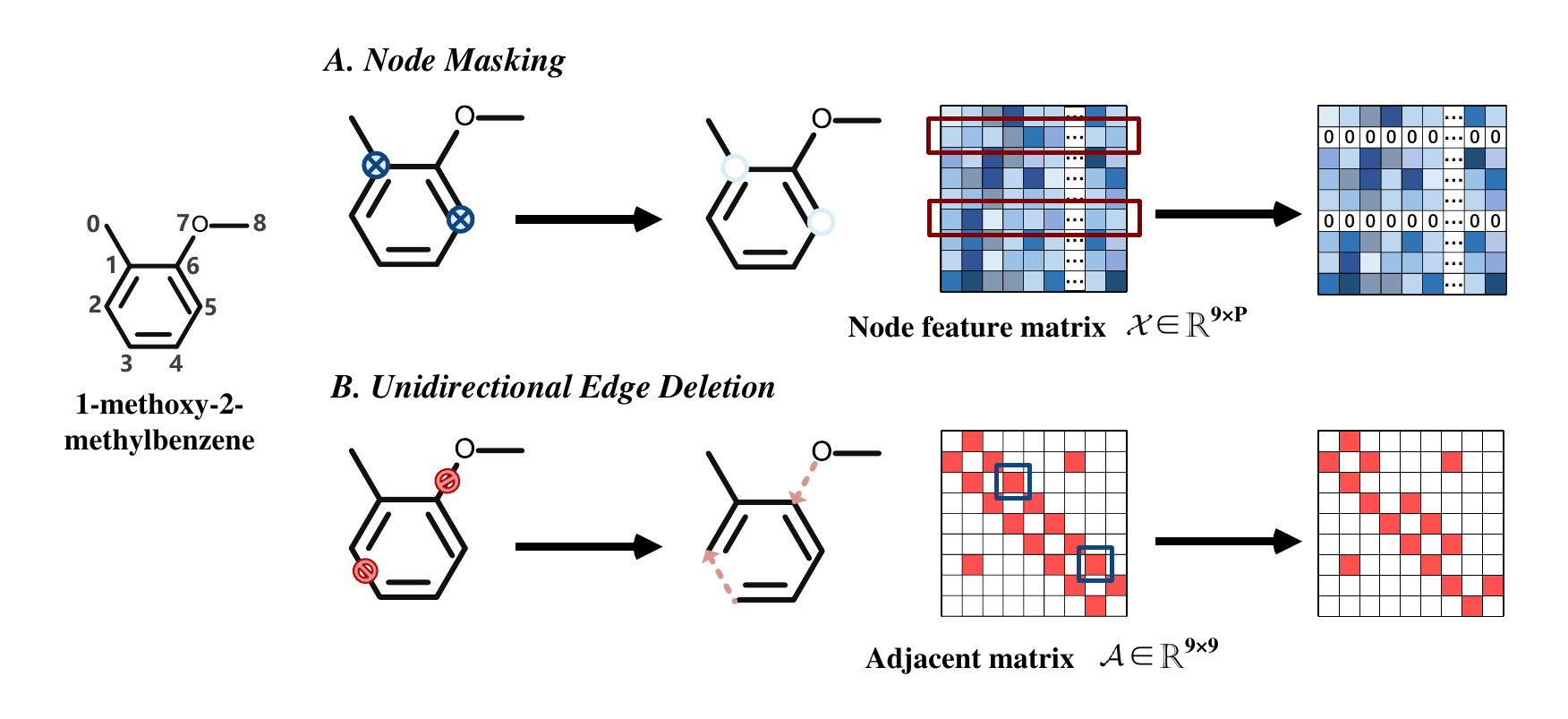}
	\caption{Graph augmentation operations with '1-methoxy-2-methylbenzene
' as the target. Node masking (Fig 2A), set the features of corresponding nodes(atoms) to zero in the feature matrix. Unidirectional edge deletion (Fig 2B), delete the one-way message passing(edges) in the adjacent matrix.}
	\label{fig:Fig 2.}
\end{figure*}

Our augmentation combines atom masking, where masked atoms have their embedding reset to 0, with unidirectional bond deletion — contrary to typical bond deletion that disrupts message-passing by severing edges. Unidirectional bond deletion only removes the connection in one direction, effectively turning an undirected bond into a directed link that maintains a one-way massage-passing during feature aggregation.
This process is represented as:

\begin{align}
\label{eq:Eq4}
\upsilon  _{i} \rightleftharpoons \upsilon_{j}\longrightarrow\left\{\begin{matrix}
\upsilon _{i} \rightharpoonup  \upsilon_{j} \\\upsilon _{i} \leftharpoondown   \upsilon_{j}
\end{matrix}\right. \;, \forall j\in \mathcal{N}(i)
\end{align}

\noindent where the symbols \begin{math}
    \rightleftharpoons ,\; \rightharpoonup
\end{math} and \begin{math}
    \leftharpoondown
\end{math} depict the nature of the link or the directionality of message passing. As for \begin{math}
    \upsilon _{i} \rightharpoonup  \upsilon_{j}
\end{math}, it implies that during the aggregation stage, \begin{math}
    \upsilon _{i} 
\end{math} is capable of transmitting its feature to its neighbor \begin{math}
   \upsilon_{j}
\end{math} along the edge, but there’s no way for it to receive feature from \begin{math}
    \upsilon _{i} 
\end{math} for updating because of the unidirectional link. Refer to Figure \ref{fig:Fig 2.}. for an intuitive demonstration of the strategy.

\subsubsection{B: Directed graph processing}

Given a graph \begin{math}
    \mathcal{G}=(\mathcal{X,A})
\end{math}, where \begin{math}
        \mathcal{X}
\end{math} and \begin{math}
        \mathcal{A}
\end{math} are node attributes and the adjacent matrix, respectively. In the adjacent matrix, \begin{math}
    a_{i,j}=1
\end{math} indicates there is an edge or link \begin{math}
    a_{i,j}=1
\end{math} between node \begin{math}
    v_{i}
\end{math} and \begin{math}
    v_{j}
\end{math} while \begin{math}
    a_{i,j}=0
\end{math} signifies the absence of direct connection between them. In the undirected graph, the edge from  \begin{math}
    v_{i}
\end{math} to \begin{math}
    v_{j}
\end{math} or the edge from \begin{math}
    v_{j}
\end{math} to \begin{math}
    v_{i}
\end{math} really represents the identical edge between them. Another way, undirected graphs are defined as directed graphs with each undirected edge represented by two directed edges. Under this definition, in the directed graph, the edge from \begin{math}
    v_{i}
\end{math} to \begin{math}
    v_{j}
\end{math} is the combination of \begin{math}
    e_{i,j}
\end{math} and \begin{math}
    e_{j,i}
\end{math}. Both of them are independent and adverse, representing the mutual influence between two nodes and serving as the route of message passing. 

When an edge in the undirected graph is deleted or blocked in one direction, the entire graph becomes directed, and the message passing along the edge becomes unidirectional and irreversible. This implies that during a message passing between two nodes, one node acts as the sender while the other must act as the receiver. Consider leveraging the centrality encoding or degree centrality, as an additional signal to the GNN encoder. The node degree is divided into two parts, out-degree and in-degree.
\begin{align}
\label{eq:Eq5}
    deg_{(v_{i})} = deg_{(v_{i})}^{+} + deg_{(v_{i})}^{-}
\end{align}
In computing centrality, if a node solely functions as a sender, it will have only out-degree; and the same is true for in-degree, only nodes that engage in both sending and receiving messages simultaneously possess out and in-degree.

\subsubsection{C: Momentum distillation}

We employ a momentum distillation pseudo-siamese network \cite{gcl, lfy2} architecture as our backbone which comprises of two networks, namely the online network and the target network. These networks map each input augmented molecular graph pair \begin{math}
    \mathcal{G}_{i}^{1}
\end{math} and \begin{math}
    \mathcal{G}_{i}^{2}
\end{math} into four slightly distinct projected vectors, denoted as \begin{math}
    \mathcal{Z}_{\theta}^1, \mathcal{Z}_{\theta}^2, \mathcal{Z}_{\xi}^1 
\end{math} and \begin{math}
    \mathcal{Z}_{\xi}^{2}
\end{math}. These four vectors represent interrelated representations of an original molecule that undergoes different combinations of augmentation operations and encoding. And the ultimate objective is to maximize the similarity between them.

Within our pseudo-siamese network architecture, both networks maintain uniformity in their fundamental structure and operational principles. Nevertheless, their delineation rests upon the functions they perform within the overarching framework. This functional divergence is principally rooted in the dissimilarity in how network parameters undergo updates during the training phase. Notably, in our framework, the target network experiences dynamic parameter updates from the online network, a departure from the conventional approach of parameter sharing or independent parameter updating processes.
\begin{align}
\label{eq:Eq6}
\theta(g,p)^{t} \longrightarrow \xi(g,p)^{t}
\end{align}
\begin{align}
\label{eq:Eq7}
\xi ^{t}=m\cdot  \xi ^{t-1} +\left ( 1-m \right ) \theta ^{t}  
\end{align}

where \begin{math}
    m
\end{math} denotes the updating rate, \begin{math}
\xi \end{math} and \begin{math}
\theta \end{math} denotes the learnable parameters of the target network and online network, respectively.

At the beginning of the training stage, the online network initializes the parameter \begin{math}
    \theta
\end{math} randomly, and the target network deep copy them to theirs \begin{math}
    \xi
\end{math}. Then the target network does not directly receive the gradient during the training, on the contrary, it updates the parameters by utilizing a momentum updating from the online network.

\subsubsection{D: Graph diffusion network}

Graph convolution is a fundamental operation for extracting node features based on their structural information. Different from the standard graph convolutional network (GCN) which employs the degree matrix to average neighboring node features, the graph diffusion network incorporates a graph diffusion process in the convolution layer \cite{diffusion}. This enhancement enables the encoder to capture the global importance and interaction effects of each node, thereby generating more informative representations.

Let a sequence of node embeddings \begin{math}
    \mathcal{G}=\left \{ \mathcal{X}_{1},\mathcal{X}_{2}\dots \mathcal{X}_{N} \right \}, \mathcal{G}\in \mathbb{R}^{N\times P}
\end{math} denotes the input graph feature, \begin{math}
\widetilde{\mathcal{A} } \in\mathbb{R}^{N\times N}    
\end{math} denotes the normalized adjacent matrix with self loops,  \begin{math}
\mathcal{H} \in\mathbb{R}^{N\times M}    
\end{math} denotes the graph embeddings, and \begin{math}
\mathcal{W} \in\mathbb{R}^{P\times M}    
\end{math} denotes the learnable model parameter matrix. The graph convolution layer is defined as
\begin{align}
\label{eq:Eq8}
    \mathcal{H}= \widetilde{\mathcal{A}}  \mathcal{X} \mathcal{W}
\end{align}

This diffusion process is characterized by a random message passing on \begin{math}
    \mathcal{G}
\end{math} with a restart probability \begin{math}
    \eta \in [0,1]
\end{math}.
\begin{align}
\label{eq:Eq9}
  \mathcal{P} = \sum_{k=0}^{\infty } \eta (1-\eta)^{k}(D_{Out}^{-1}W)^{k}  
\end{align}

Where \begin{math}
    k
\end{math} is the step of diffusing, and \begin{math}
    D_{Out}^{-1}W
\end{math} indicates the state transition matrix which indicates paths for message passing. And \begin{math}
    D_{Out} = diag(W1)
\end{math} is the out-degree diagonal matrix with all one's entries. Since the diffusion process exhibits Markovian properties, \begin{math}
    \mathcal{P}\in \mathbb{R} ^{N\times N}
\end{math} converges to a stationary distribution after a finite number of steps \cite{dcrnn}. And each row in \begin{math}
    \mathcal{P}
\end{math}, \begin{math}
    \mathcal{P}[i,:]\in\mathbb{R} ^{N}
\end{math} represents the possibility of diffusion from \begin{math}
    v_{i}
\end{math} to its neighbors.

The diffusion convolution operation over a graph input embeddings 
\begin{math}
    \mathcal{G}\in \mathbb{R}^{N\times P}
\end{math} and a filter \begin{math}
    f_{\phi } 
\end{math} is defined as:
\begin{align}
\label{eq:Eq10}
    \mathcal{X} f_{\phi } = \sum_{k=0}^{K-1} (\phi_{k} (D_{Out}^{-1}W)^{k})\mathcal{X}
\end{align}
where \begin{math}
    \mathcal{\phi}\in \mathbb{R}^{N}
\end{math} are the parameters for the filter and \begin{math}
    D_{Out}^{-1}W
\end{math} represents the transition matrices of the diffusion process.

However, the computation of diffusion processes is both resource-intensive and time-consuming. We generalized and simplified the diffusion process. The diffusion graph convolution layer with \begin{math}
    K
\end{math} finite steps is written as 

\begin{align}
\label{eq:Eq11}
\mathcal{H} =\sigma ( \sum_{k = 0}^{K} \mathcal{P}^{k} \mathcal{X}W_{k}) 
\end{align}
where \begin{math}
    \mathcal{P}^{k}
\end{math} represents the power series of the transition matrix, it is equivalent to \begin{math}
    (D_{Out}^{-1}\mathcal{W})^{k}
\end{math} in the equation before simplification, and \begin{math}
\mathcal{P} = \mathcal{A}/rowsum(\mathcal{A})     
\end{math}. \begin{math}
    \mathcal{W}\in\mathbb{R}^{P\times M} 
\end{math} denotes the learnable matrix, and \begin{math}
    \mathcal{H}\in\mathbb{R}^{N\times B}
\end{math} denotes the output graph embeddings, and \begin{math}
    \sigma(.)
\end{math} is the activation function (e.g., ReLU and Sigmoid).
  
To capture comprehensive structural information within the graph, we incorporate a bidirectional diffusion mechanism by adding a reversed-direction diffusion process. 
\begin{align}
\label{eq:Eq12}
     \mathcal{H}=\sigma (\sum_{k=0}^{K}(\epsilon  \mathcal{P}_{f}^{k}\mathcal{X}\mathcal{W}_{k1} + (1-\epsilon )\mathcal{P}_{f}^{b}\mathcal{X}\mathcal{W}_{k2} ) ) 
\end{align}
where \begin{math}
    \epsilon
\end{math} denotes a constant coefficient. Under this configuration, the diffusion process has two directions, the forward and backward directions. And the transition matrix \begin{math}
    \mathcal{P}
\end{math} is divided into two components, the forward transition matrix \begin{math}
    \mathcal{P}_{f} = \mathcal{A}/rowsum(\mathcal{A})
\end{math} and the backward transition matrix \begin{math}
    \mathcal{P}_{b} = \mathcal{A}^{T}/rowsum(\mathcal{A}^{T})
\end{math}. It is worth noting, that these transition matrices also reflect the out-degree and in-degree distribution of nodes and the node centrality. For one  node \begin{math}
    v_{i}
\end{math} within the graph, \begin{math}
    deg_{v_{i}}^{+} = rowsum(\mathcal{A})_{i}
\end{math}, similarly, \begin{math}
    deg_{v_{i}}^{-} = rowsum(\mathcal{A}^{T})_{i}
\end{math}

\subsubsection{E: Dual-interaction}
Following the application of various graph augmentation operations and momentum distillation diffusion graph networks, the online graph encoder and target graph encoder \begin{math}
    g_{\theta}
\end{math} and \begin{math}
    g_{\xi}\end{math} both map \begin{math}
    \mathcal{G}_{i}^{1}
\end{math} and \begin{math}
    \mathcal{G}_{i}^{2}
\end{math} to a feature embedding space. 
To facilitate contrastive loss computation, the resulting feature vectors are projected onto an invariant space \begin{math}
    \mathbb{R}^{d}
\end{math} using either projection head \begin{math}
    p_{\theta}
\end{math} or \begin{math}
    p_{\xi}
\end{math}.
Finally, we obtained four interconnected vectors \begin{math}
    \mathcal{Z}_{\theta}^{1}, \mathcal{Z}_{\theta}^{2}, \mathcal{Z}_{\xi}^{1} \end{math} and  
 \begin{math}
     \mathcal{Z}_{\xi}^{2}
\end{math}, which are derived from an original molecule graph \begin{math}
    \mathcal{G}_{i}
\end{math}. However, when utilizing them to calculate the contrastive loss, they can be categorized into the following three distinct scenarios.

The objective of training for graph-interaction instance discrimination is to maximize the similarity between projection vectors originating from distinct augmented graphs but embedded by the same network. Specifically, is the similarity between \begin{math}
    \mathcal{Z}_{\theta}^{1} \;and\; \mathcal{Z}_{\theta}^{2} \end{math}, and \begin{math}
 \mathcal{Z}_{\xi}^{1} \;and\;
     \mathcal{Z}_{\xi}^{2}
\end{math}. Meanwhile, minimize the similarity with projected vectors of all the other molecular graphs in the same mini-batch. Take \begin{math}
\mathcal{Z}_{\theta}^{1} \;and\; \mathcal{Z}_{\theta}^{2} \end{math} as example, \begin{math}
\mathcal{Z}_{\theta}^{1}=p_{\theta}(g_{\theta}(\mathcal{G}^{1})),\;  \mathcal{Z}_{\theta}^{2}=p_{\theta}(g_{\theta}(\mathcal{G}^{2}))     
\end{math} they are the final vectors obtained through online graph encoder \begin{math}
        g_{\theta}
\end{math} and online projector \begin{math}
        p_{\theta}
\end{math}, meanwhile \begin{math}
        \mathcal{G}^{1}
\end{math} and \begin{math} \mathcal{G}^{2} \end{math} are derived from an identical molecular graph. Next,the similarity \begin{math}
    sim(\mathcal{Z}_{\theta}^{1}, \mathcal{Z}_{\theta}^{2})
\end{math} between two projected vectors is computed. 
\begin{align}
\label{eq:Eq13}
    sim(\mathcal{Z}_{\theta}^{1}, \mathcal{Z}_{\theta}^{2}) = \frac{\mathcal{Z}_{\theta}^{1}\cdot \mathcal{Z}_{\theta}^{2}}{\left \| \mathcal{Z}_{\theta}^{1} \right \| \left \| \mathcal{Z}_{\theta}^{2} \right \| } 
\end{align}

We adopted the normalized temperature-scaled cross entropy (NT-Xent) \cite{conloss} as the contrastive loss function, 
\begin{align}
\label{eq:Eq14}
    \mathcal{L}_{GI}= \frac{exp(sim(\mathcal{Z}_{\theta}^{1}, \mathcal{Z}_{\theta}^{2})/\tau)}{ { \sum_{b=1}^{2B}\mathbf{1}_{[b\not{=}1]}exp(sim(\mathcal{Z}_{\theta}^{1}, \mathcal{Z}_{b})/\tau)}}  
\end{align}
where \begin{math}
    \mathbf{1}_{[b\not{=}1]\in[0,1]}
\end{math} is an indicator function evaluating to 1 iff \begin{math} b \neq 1
\end{math}, \begin{math}
    \tau
\end{math} denotes a temperature parameter; \begin{math}
    B
\end{math} is the number of samples in a mini-batch. The loss between \begin{math}
 \mathcal{Z}_{\xi}^{1} \;and\;
     \mathcal{Z}_{\xi}^{2}
\end{math} is computed similar.

Different from the graph-interaction, encoder-interaction instance discrimination seeks to maximize the similarity between projection vectors obtained from the same augmented graphs but embedded by different networks. In essence, this involves assessing the similarity between representations derived from a single augmented graph and embedded by both the online network and the target network, respectively. Such as \begin{math}
 \mathcal{Z}_{\theta}^{1} \;and\;
     \mathcal{Z}_{\xi}^{1}
\end{math}, and \begin{math}
 \mathcal{Z}_{\theta}^{2} \;and\;
     \mathcal{Z}_{\xi}^{2}
\end{math}. And the cross entropy loss \begin{math}
    \mathcal{L}_{EI}
\end{math} exhibits similarities with \begin{math}
    \mathcal{L}_{GI}
\end{math}.

According to the rule of pairwise interactions, four projection vectors \begin{math}
    \mathcal{Z}_{\theta}^{1}, \mathcal{Z}_{\theta}^{2}, \mathcal{Z}_{\xi}^{1} \end{math} and  
 \begin{math}
     \mathcal{Z}_{\xi}^{2}
\end{math}, can generate a total of three distinct types of six interactions. Apart from the graph-interaction and encoder-interaction, a new form of interaction emerges between \begin{math}
 \mathcal{Z}_{\theta}^{1} \;and\;
     \mathcal{Z}_{\xi}^{2}
\end{math}, and \begin{math}
 \mathcal{Z}_{\theta}^{2} \;and\;
     \mathcal{Z}_{\xi}^{1}
\end{math}. These two projection vectors originate from different augmented graphs and networks, making this interaction a fusion of graph-interaction and encoder-interaction, termed "multi-interaction". And the cross entropy loss \begin{math}
    \mathcal{L}_{MI}
\end{math} is also calculated in the same manner.

Joint loss \begin{math}\mathcal{L}_{Joint}\end{math} work as the overall training objective in the optimization process, which consists of three parts: \begin{math}
    \mathcal{L}_{GI},\mathcal{L}_{EI}
\end{math} and  \begin{math}
    \mathcal{L}_{MI}
\end{math}. 
\begin{align}
\label{eq:Eq17}
    \mathcal{L} _{\small {Joint}} = \alpha\mathcal{L} _{\small {GI}} + \beta \mathcal{L} _{\small {EI}} + \gamma \mathcal{L} _{\small {MI}}
\end{align}

Where \begin{math}
    \alpha, \beta
\end{math} and \begin{math}
    \gamma
\end{math} represent three independent constant coefficients.

\section{Results and discussion}

\subsection{Experimental setting}

\subsubsection{Fine-tune and downstream tasks}

In the pre-training stage, we employed batch gradient descent to minimize the contrastive loss. Moving to the transfer learning phase for molecular property prediction, DIG-Mol involved integrating two fully connected layers following the GNN encoder. During the fine-tuning phase, we fixed the weights of the GNN encoder and exclusively fine-tuned the parameters of the two additional fully connected layers.

\paragraph{\textbf{Downstream benchmarks}}

We assessed the performance of DIG-Mol using a diverse suite of 13 datasets from MoleculeNet \cite{moleculenet}, covering various domains, including physiology, biophysics, physical chemistry, and quantum mechanics. Each dataset focuses on different molecule properties. These datasets fall into two categories based on their task requirements: classification and regression tasks. Basic details of these datasets are summarized in Table \ref{tab:Tab 1.}, while detailed information is provided in Appendix \textup{IV}, Table \ref{tab:Atab 2.}.

\begin{table}[htb]
    \centering
    \caption{Brief induction of downstream tasks used in this work.}
    \label{tab:Tab 1.}
    \begin{tabular}{lllll}
    \hline
        Dataset & \#Molecule & \#Task & Task type & Metric  \\ \hline
        BBBP & 2039 & 1 & Classification  & ROC-AUC    \\ 
        BACE & 1513 & 1 & Classification  & ROC-AUC    \\ 
        SIDER & 1427 & 27 & Classification  & ROC-AUC    \\ 
        ClinTox & 1478 & 2 & Classification  & ROC-AUC    \\ 
        HIV & 41127 & 1 & Classification  & ROC-AUC    \\ 
        Tox21 & 7831 & 12 & Classification  & ROC-AUC\\
        MUV & 93087 & 17 & Classification & PRC-AUC
        \\ \hline
        FreeSolv  & 642 & 1 & Regression & RMSE  \\ 
        ESOL & 1128 & 1 & Regression & RMSE  \\ 
        Lipo & 4200 & 1 & Regression & RMSE  \\ 
        QM7 & 6830 & 1 & Regression & MAE  \\ 
        QM8 & 21786 & 12 & Regression & MAE \\ 
        QM9 & 130829 & 8 & Regression & MAE\\ \hline
    \end{tabular}
    
\end{table}

\paragraph{\textbf{Few shot tasks}}

To demonstrate the robust generalization and adaptability of DIG-Mol, we conducted a series of few-shot experiments using the multi-task datasets Tox21 and SIDER. These datasets were specifically chosen for their suitability in few-shot task scenarios. Detailed information about these benchmark datasets is available in Table \ref{tab:Tab 2.}.

\begin{table}[htb]
    \centering
    \caption{Configurations of few-shot tests.}
    \label{tab:Tab 2.}
    \setlength{\tabcolsep}{6mm}{}
    \begin{tabular}{lcc}
    \hline
        Dataset & Tox21 & SIDER  \\ \hline
        \#Molecule & 7831 & 1427   \\ 
        \#Tasks & 12 & 27  \\ 
        \#Training tasks & 9 & 21 \\ 
        \#Testing  tasks & 3 & 6     \\ \hline   
    \end{tabular}
\end{table}

\subsubsection{Baselines}

We conducted a comparative analysis of DIG-Mol’s performance against various representative and cutting-edge baseline methods. These methodologies are broadly classified into two groups based on their learning paradigms: supervised and self-supervised learning methods. In the realm of supervised learning, SchNet \cite{schnet} builds upon Deep Transition Neural Networks, effectively incorporating molecular information. MGCN \cite{mgcn} represents Multi-level Graph Neural Networks and utilizes a hierarchical approach in the context of GNNS. D-MPNN \cite{dmpnn} is notable for its use of directed message-passing mechanisms. In the self-supervised learning category, N-Gram \cite{ngarm} and Hu et al. \cite{hu} combine multiple short walks and multi-level information within GNN. Additionally, MolCLR \cite{nat1} stands out as an efficient contrastive learning framework using GCN or GIN as graph encoders. Detailed information on all the baseline models is provided in Appendix \textup{VI}.

\subsubsection{Parameter setting}

In training DIG-Mol, we employed the Adam optimizer during both pre-training and fine-tuning phases, beginning with an initial learning rate of 0.001. We integrated a cosine annealing schedule to adjust the learning rate over 100 training epochs. For the pre-training phase, our architecture is centered around a five-layer GNN, with the temperature \begin{math}
    \tau 
\end{math} set to 0.1 in the contrastive loss function to optimize the learning process. After pretraining, we remove the target graph encoder \begin{math}
g_{\xi}
\end{math} and both projection heads \begin{math}
    p_{\theta}
\end{math} and \begin{math}
    p_{\xi}
\end{math}, proceeding solely with the pre-trained online graph encoder \begin{math}
    g_{\theta}
\end{math} and a new projection head \begin{math}
    p_{f}
\end{math} for fine-tuning tasks.

The fine-tuning phase is characterized by its adaptable configurations, as we conducted a random search for the best combination of hyper-parameters to ensure the model achieves its optimal performance. The varied hyper-parameters and their specific configurations explored during this phase are documented in detail in Appendix \textup{V}, Table \ref{tab:Atab 3.}.

\subsubsection{Computing infrastructures}

The DIG-Mol framework was implemented using PyTorch version 1.11.0. The experimental tasks were performed on a workstation equipped with dual NVIDIA GeForce RTX 3090 GPUs, each boasting 24GB of memory. The system includes an Intel Core i9-10900x (3.70 GHz) CPU supported by 62.5 GB of RAM.

\subsection{DIG-Mol achieved state-of-the-art performance}

In our comprehensive experimental evaluation, we measured the performance of DIG-Mol against ten leading baseline models on a range of classification and regression tasks. The comparative results, including test ROC-AUC (\%) and PRC-AUC (\%) for classification tasks, are detailed in Table \ref{tab:Tab 3.}. For regression tasks, where different metrics were appropriate, we measured FreeSolv, ESOL, and Lipo using RMSE and QM7, QM8, and QM9 using MAE; these results are summarized in Table \ref{tab:Tab 4.}. To confirm the robustness of our findings, we report both the mean and standard deviation from three separate runs for each task.

\begin{table*}[!ht]
    \centering
    \caption{Test performance of different models across seven classification benchmarks. The first seven models are supervised
learning methods and the last four are self-supervised methods. The mean and standard deviation of test ROC-AUC or PRC-AUC(for MUV) (\%) on each benchmark are reported.
}
    \label{tab:Tab 3.}
    \setlength{\tabcolsep}{3.7mm}{}
    \begin{tabular}{lllllllll}
    \hline
        Dataset & BBBP & BACE & SIDER & ClinTox & HIV & Tox21 & MUV \\ \hline
        \#Molecules & 2039 & 1513 & 1427 & 1478 & 41127 & 7831 & 93087\\ 
        \#Tasks & 1 & 1 & 27 & 2 & 1 & 12 & 17\\ \hline
        RF  & 71.4 ± 0.0 & 86.7 ± 0.8 & \begin{math}
            \mathbf{68.4\pm0.9}
        \end{math} & 71.3 ± 5.6 & 78.1 ± 0.6 & 76.9 ± 1.5 & 63.2 ± 2.3\\ 
        SVM & 72.9 ± 0.0 & 86.2 ± 0.0 & 68.2 ± 1.3 & 66.9 ± 9.2 & \begin{math}
            \mathbf{79.2\pm0.0}
        \end{math} & \begin{math}
            \mathbf{81.8\pm1.0}
        \end{math} & 67.3 ± 1.3\\ 
        GCN & 71.8 ± 0.9 & 71.6 ± 2.0 & 53.6 ± 3.2 & 62.5 ± 2.8 & 74.0 ± 3.0 & 70.9 ± 2.6 & 71.6 ± 4.0\\ 
        GIN  & 65.8 ± 4.5 & 70.1 ± 5.4 & 57.3 ± 1.6 & 58.0 ± 4.4 & 75.3 ± 1.9 & 74.0 ± 0.8 & 71.8 ± 2.5\\ 
        SchNet & 84.8 ± 2.2 & 76.6 ± 1.1 & 53.9 ± 3.7 & 71.5 ± 3.7 & 70.2 ± 3.4 & 77.2 ± 2.3 & 71.3 ± 3.0\\ 
        MGCN & \begin{math}
            \mathbf{85.0\pm6.4}
        \end{math} & 73.4 ± 3.0 & 55.2 ± 1.8 & 63.4 ± 4.2 & 73.8 ± 1.6 & 70.7 ± 1.6 & 70.2 ± 3.4\\ 
        D-MPNN & 71.2 ± 3.8 & \begin{math}
            \mathbf{85.3\pm5.3}
        \end{math} & 63.2 ± 2.3 & \begin{math}
            \mathbf{90.5\pm5.3}
        \end{math} & 75.0 ± 2.1 & 68.9 ± 1.3 &  76.2 ± 2.8\\ \hline
        N-Gram & \begin{math}
            \mathbf{91.2\pm3.0}
        \end{math} & 87.6 ± 3.5 & 63.2 ± 0.5 & 85.5 ± 3.7 & \begin{math}
            \mathbf{83.0\pm1.3}
        \end{math} & 76.9 ± 2.7 & 81.6 ± 1.9\\ 
        Hu et.al & 70.8 ± 1.5 & 85.9 ± 0.8 & 65.2 ± 0.9 & 78.9 ± 2.4 & 80.2 ± 0.9 & 78.7 ± 0.4 & 81.4 ± 2.0\\ 
        MolCLR & 73.8 ± 0.2  & \underline{89.0 ± 0.3} & \underline{68.0 ± 1.1} & \underline{93.2 ± 1.7} & 80.6 ± 1.1 & \underline{79.8 ± 0.7} & \underline{88.6 ± 1.2}\\ 
        DIG-Mol & \underline{77.0 ± 0.5} & \begin{math}
            \mathbf{91.2\pm0.3}
        \end{math} & \begin{math}
            \mathbf{71.8\pm0.4}
        \end{math} & \begin{math}
            \mathbf{94.6\pm1.1}
        \end{math} & \underline{81.3 ± 0.3}  & \begin{math}
            \mathbf{80.1\pm0.9}
        \end{math} & \begin{math}
            \mathbf{89.1\pm1.5}
        \end{math} \\  \hline
    \end{tabular}
    \raggedright
    *The best performing supervised and self-supervised methods for each benchmark are marked as bold, while the second-best are underlined.
\end{table*}

\begin{table*}[!ht]
    \centering
    \caption{Test performance of different models across six regression benchmarks. The first seven models are supervised
learning methods and the last four are self-supervised methods. Mean and standard deviation of test RMSE (for FreeSolv, ESOL, Lipo) or MAE (for QM7, QM8, QM9) are reported.}
    \label{tab:Tab 4.}
    \setlength{\tabcolsep}{3.6mm}{}
    \begin{tabular}{lllllll}
    \hline
        Dataset & FreeSolv  & ESOL & Lipo & QM7 & QM8 & QM9 \\ \hline
        \#Molecules & 642 & 1128 & 4200 & 6830 & 21786 & 130829 \\ 
        \#Tasks & 1 & 1 & 1 & 1 & 12 & 8 \\ \hline
        RF  & \begin{math}
            \mathbf{2.03\pm0.22}
        \end{math} & 1.07 ± 0.19 & 0.88 ± 0.04 & 122.7 ± 4.2 & 0.0423 ± 0.0021 & 16.061 ± 0.019 \\ SVM & 3.14 ± 0.00 & 1.50 ± 0.00 & 0.82 ± 0.00 & 156.9 ± 0.0 & 0.0543 ± 0.0010 & 24.613 ± 0.144 \\ 
        GCN & 2.87 ± 0.14 & 1.43 ± 0.05 & 0.85 ± 0.08 & 122.9 ± 2.2 & 0.0366 ± 0.0011 & 5.796 ± 1.969 \\ 
        GIN  & 2.76 ± 0.18 & 1.45 ± 0.02 & 0.85 ± 0.07 & 124.8 ± 0.7 & 0.0371 ± 0.0009 & 4.741 ± 0.912 \\
        SchNet & 3.22 ± 0.76 & 1.05 ± 0.06 & 0.91 ± 0.10 & \begin{math}
            \mathbf{74.2\pm6.0}
        \end{math} & 0.0204 ± 0.0021 & 0.081 ± 0.001 \\ 
        MGCN & 3.35 ± 0.01 & 1.27 ± 0.15 & 1.11 ± 0.04 & 77.6 ± 4.7 & 0.0223 ± 0.0021 & \begin{math}
            \mathbf{0.050\pm0.002}
        \end{math} \\ 
        D-MPNN & 2.18 ± 0.91 & \begin{math}
            \mathbf{0.98\pm0.26}
        \end{math} & \begin{math}
            \mathbf{0.65\pm0.05}
        \end{math} & 105.8 ± 13.2 & \begin{math}
            \mathbf{0.0143\pm0.0022}
        \end{math} & 3.241 ± 0.119 \\ \hline
        N-Gram & 2.51 ± 0.19 & \underline{1.10 ± 0.03} & 0.88 ± 0.12 & 125.6 ± 1.5 & 0.0191 ± 0.0003 & 7.636 ± 0.027 \\ 
        Hu et.al & 2.83 ± 0.12 & 1.22 ± 0.02 & \underline{0.740 ± 0.00} & 110.2 ± 6.4 & 0.0320 ± 0.0032 & 4.349 ± 0.061 \\ 
        MolCLR & \underline{2.20 ± 0.20} & 1.11 ± 0.01 & 
        \begin{math}
            \mathbf{0.65\pm0.08}
        \end{math} & \underline{83.1 ± 4.0} &  \underline{0.0174 ± 0.0013} & \begin{math}
            \mathbf{2.357\pm0.118}
        \end{math} \\ 
        DIG-Mol &\begin{math}
            \mathbf{2.05\pm0.15}
        \end{math} & \begin{math}
            \mathbf{1.09\pm0.02}
        \end{math} & \begin{math}
            \mathbf{0.65\pm0.04}
        \end{math} & \begin{math}
            \mathbf{73.9\pm3.5}
        \end{math} & \begin{math}
            \mathbf{0.0172\pm0.0021}
        \end{math} & \underline{3.743 ± 0.125} \\ \hline
    \end{tabular}
    \raggedright
    *The best performing supervised and self-supervised methods for each benchmark are marked as bold, while the second-best are underlined.
\end{table*}

 \subsubsection{Classification tasks}
Several noteworthy points can be drawn from Table \ref{tab:Atab 2.}.:

(1) DIG-Mol outperforms other self-supervised learning models across most benchmarks, including BACE, SIDER, ClinTox, Tox21 and MUV. This underscores the superior efficacy of the DIG-Mol framework within the realm of self-supervised learning strategy.

(2) DIG-Mol also maintains a strong competitive edge when measured against the top-performing supervised learning baselines, showcasing its consistent and robust performance across different learning paradigms. In some cases, DIG-Mol has exceeded the performance of leading supervised learning methods, which often rely on sophisticated aggregation techniques and specialized feature engineering.

(3) DIG-Mol notably performs well with datasets featuring a limited number of molecules, like ClinTox, BACE, and SIDER, demonstrating its proficiency in learning transferable and significant representations. However, its predictive performance on the BBBP and HIV datasets falls short of the best. This is likely due to the distinct natures of these datasets: the BBBP’s molecular properties are highly sensitive to structural nuances, and the HIV dataset suffers from a significant imbalance in its molecular distribution. These factors present complex challenges that hinder any model’s ability to excel universally across all task types without tailored designs.

Additionally, Figure \ref{fig:Fig 3.}. showcases the Receiver Operating Characteristic Curves (ROC) curves for various datasets. Notably, the SIDER dataset’s results align closely with those of Tox21, and to maintain brevity, we have excluded them from the figure.

\begin{figure*}[htb]
	\centering
	\begin{minipage}{0.32\linewidth}
		\centering
		\includegraphics[width=1\linewidth]{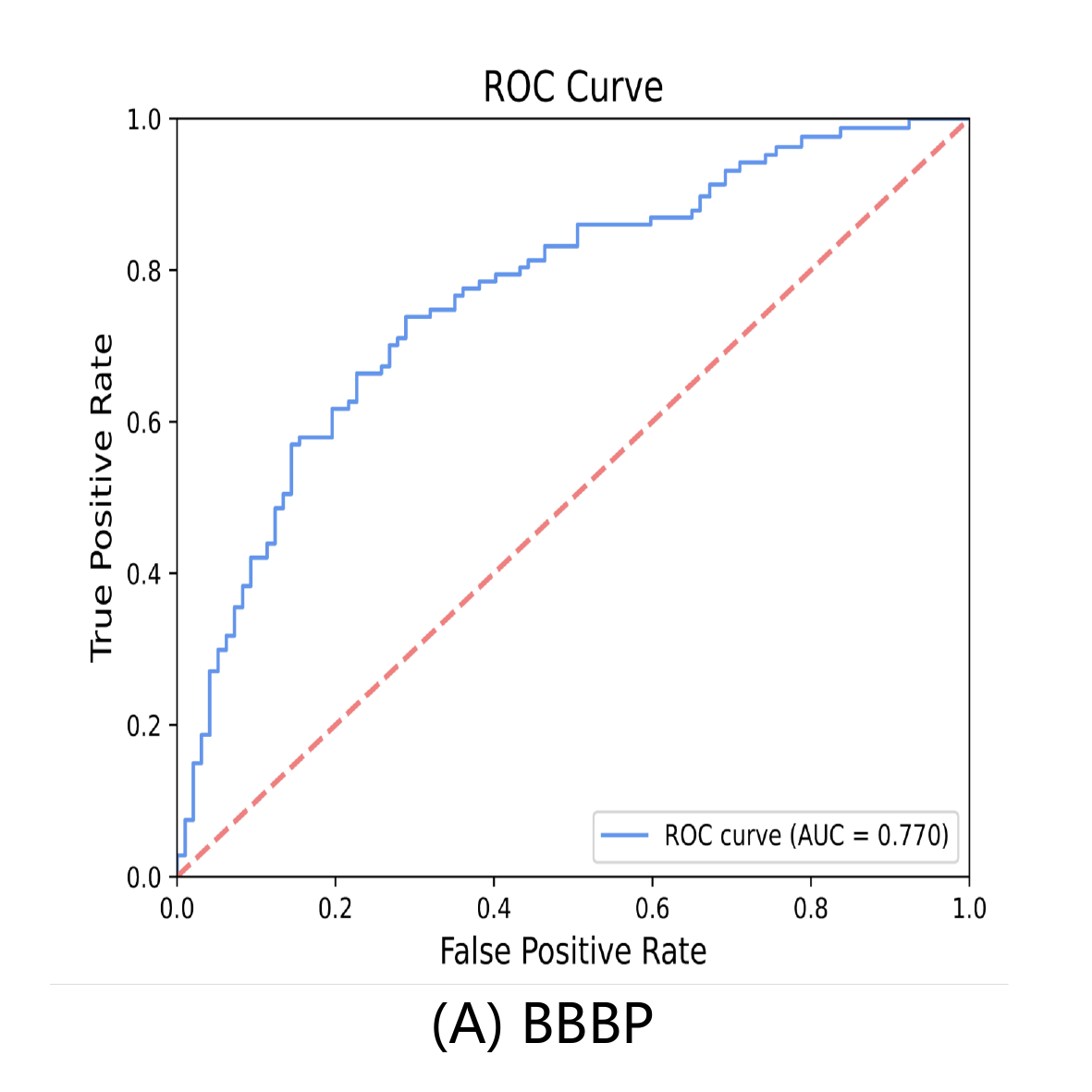}
	\end{minipage}
	\begin{minipage}{0.32\linewidth}
		\centering
		\includegraphics[width=1\linewidth]{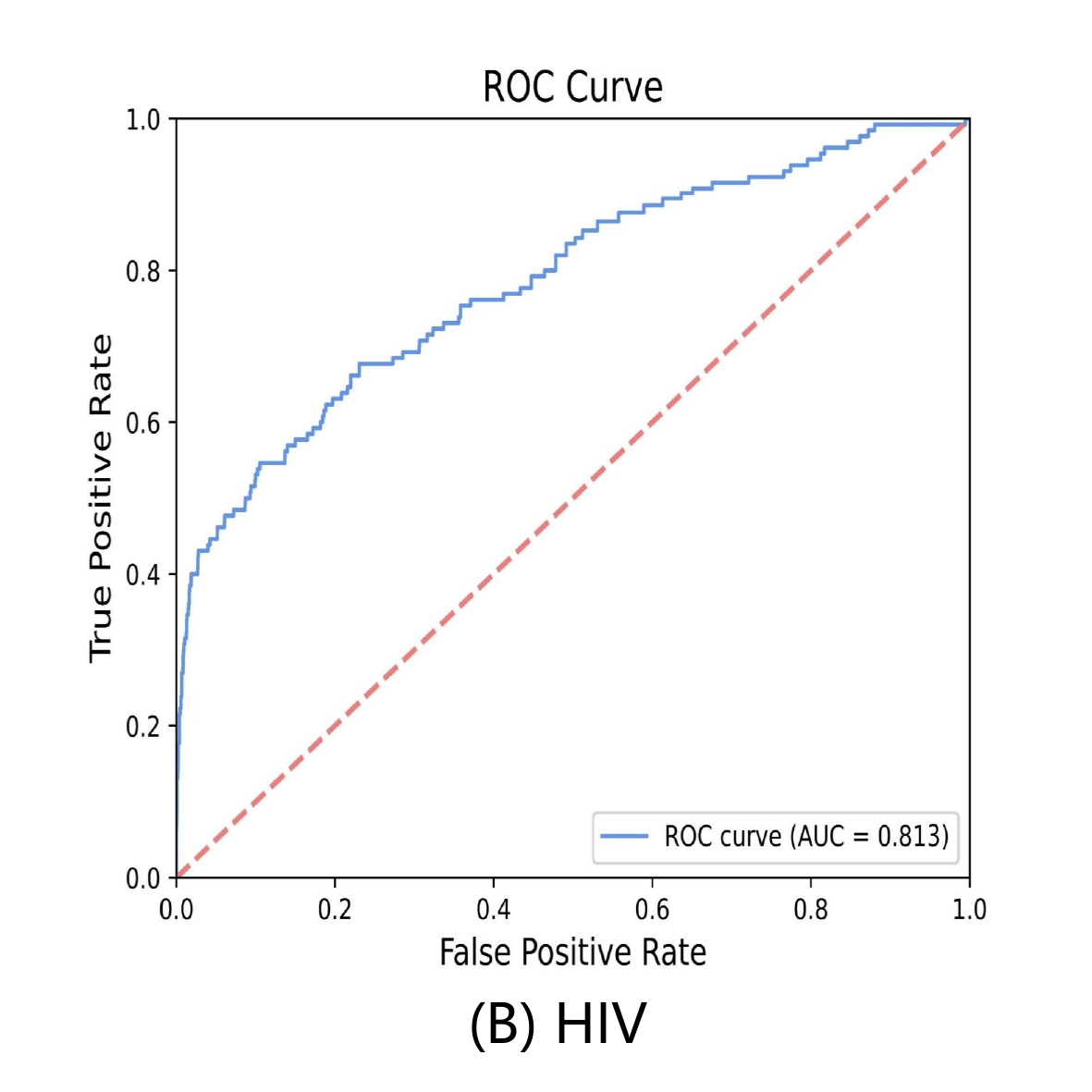}
	\end{minipage}
        \begin{minipage}{0.32\linewidth}
		\centering
		\includegraphics[width=1\linewidth]{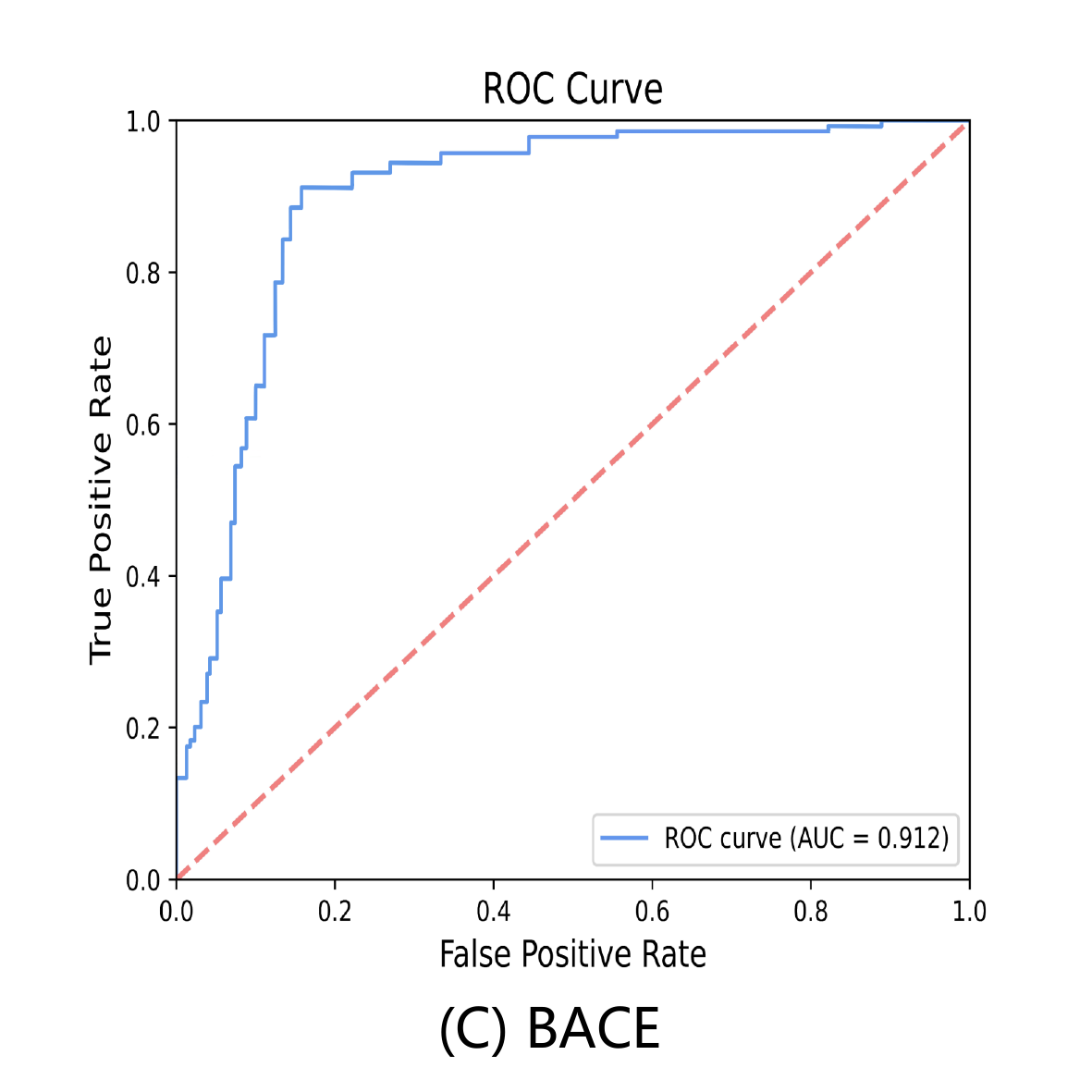}
	\end{minipage}

 \vspace{-5pt}
 \centering
	\begin{minipage}{0.32\linewidth}
		\centering
		\includegraphics[width=1\linewidth]{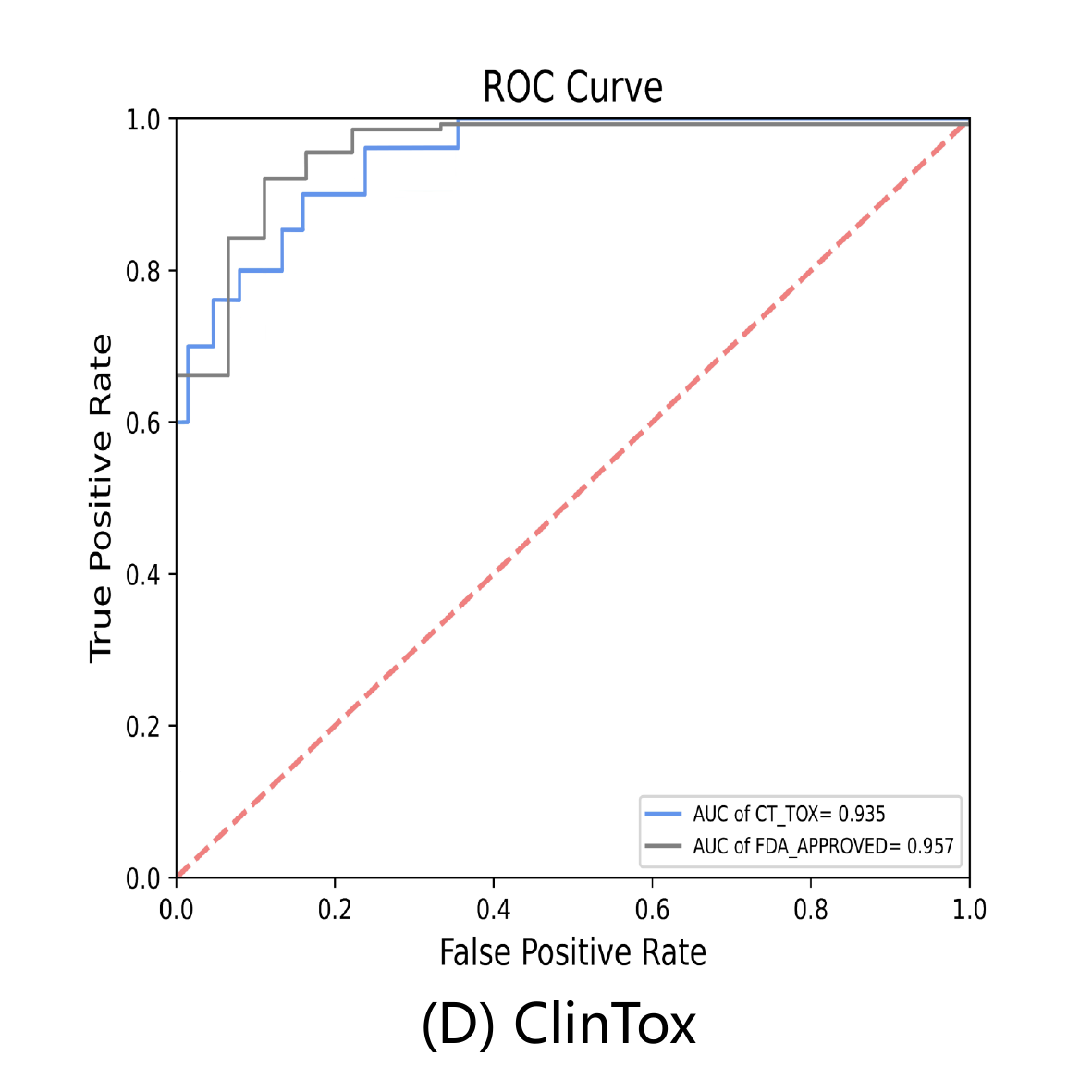}
	\end{minipage}
	\begin{minipage}{0.32\linewidth}
		\centering
		\includegraphics[width=1\linewidth]{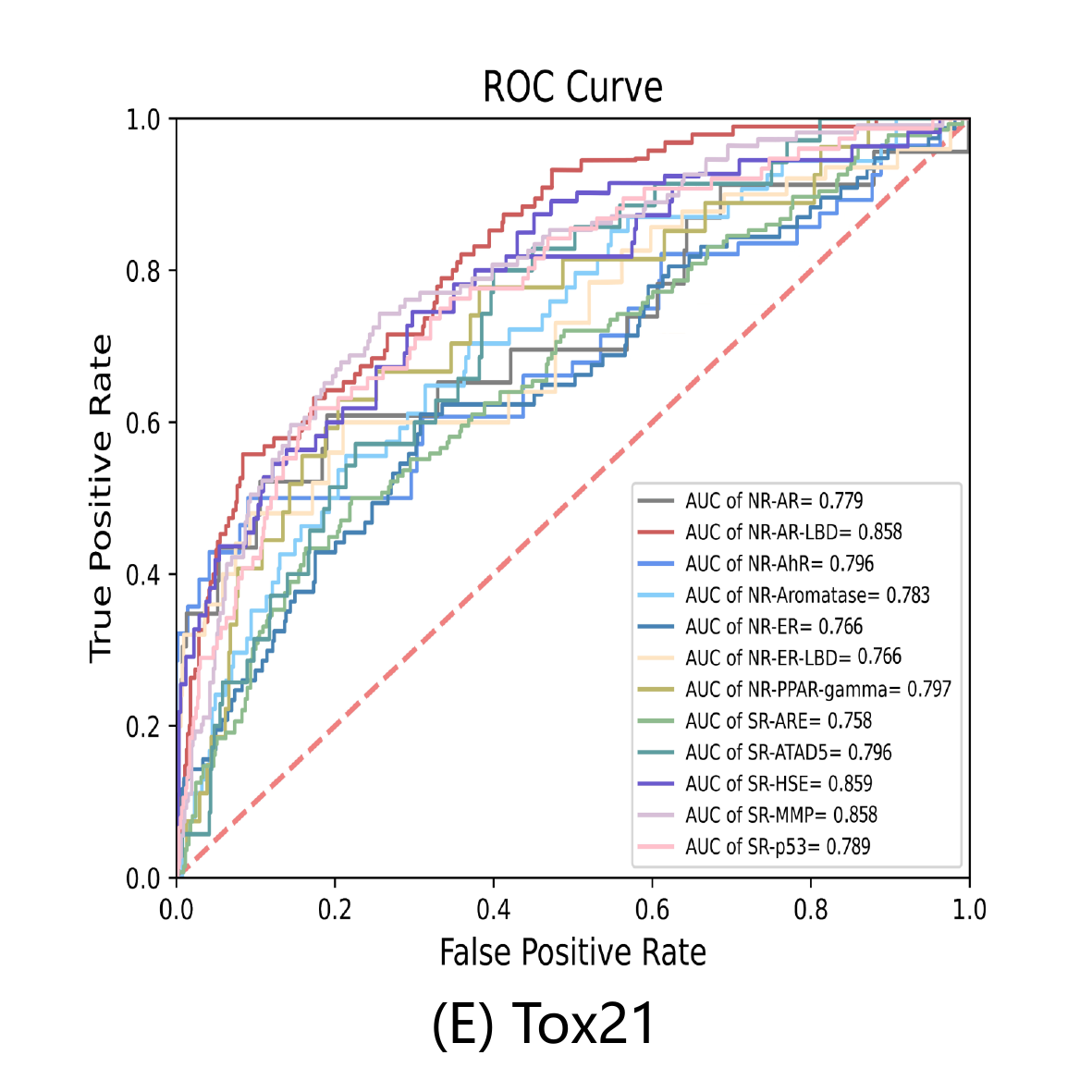}
	\end{minipage}
\end{figure*}

\begin{figure*}
    \centering
    \caption{The ROC curves for different classification datasets. The different colored curves represent different subtasks. Please note that the ROC-AUCs plot may not exactly correspond to the best results in the figure legends.}
    \label{fig:Fig 3.}
\end{figure*}

\subsubsection{Regression tasks}

In the realm of regression tasks, which inherently present greater challenges than classification tasks due to their reliance on manually defined discrete labels. In spite of this, DIG-Mol still showcases notable performance. Table \ref{tab:Tab 4.}. reveals:

(1) Across all regression benchmarks, DIG-Mol generally outperformed self-supervised baselines, with QM9 being a tight competitive where it still closely approaches the top results. Its performance on datasets like FreeSolv, Lipo, QM7 and QM8 is remarkably competitive and, in some cases, superior to supervised baselines.

(2) Compared to MolCLR, which utilizes a similar GCL (graph contrastive learning) framework and graph encoders (GCN and GIN), DIG-Mol exhibits a significant performance boost across nearly all datasets. Notably, this is achieved with a smaller input batch size during pretraining, a deliberate choice to enhance reproducibility and conserve computational resources. The superior performance of DIG-Mol suggests that its unique design features, such as the momentum distillation and dual-interaction mechanisms, are effective in capturing critical molecular representations during pretraining, thereby bolstering the overall efficacy of contrastive learning approach.

(3) Supervised learning models, particularly on regression task datasets, tend to achieve high performance, often out of reach for many self-supervised learning models. For example, MGCN and SchNet excel on the QM9 dataset due to their specialized design for quantum interactions and the inclusion of 3D positional data. These features enable them to perform exceptionally well on datasets aimed at evaluating quantum mechanical properties of molecules. However, their specialized nature limits their versatility, making them less competitive across other benchmarks. In contrast, DIG-Mol is more comprehensive in its performance.

\subsection{Few shot tasks}

In the drug discovery pipeline, where only a selected group of candidate molecules are further synthesized and tested in the subsequent wet-lab conditions, models should efficiently learn from large datasets and adeptly transfer this knowledge to tasks with very few samples. To demonstrate the capabilities of our model, DIG-Mol, in such few-shot learning scenarios, we conducted experiments on the Tox21 and SIDER datasets, benchmarking against popular GNN variant models, including GCN \cite{gcn1}, GIN \cite{ginfew}, and GAT \cite{gat2}. The specific results are detailed in Table 5, show DIG-Mol’s superior performance on both Tox21 and SIDER datasets. This success can be attributed to the rich and informative molecular representations learned by DIG-Mol during its contrastive learning phase, which employs a dual-interaction mechanism. These representations allow DIG-Mol to quickly adapt to new tasks with sparse data. It's important to note that while DIG-Mol was not specifically designed for few-shot tasks, its strong performance in these tests underscores its potential for effective knowledge transfer in data-constrained scenarios. Nonetheless, DIG-Mol is not intended to outperform models that are specialized for few-shot learning.

\begin{table*}[!ht]
\centering
\caption{Few-shot test performance of different models across Tox21 and SIDER datasets.}
\label{tab:Tab 5.}
\setlength{\tabcolsep}{6mm}{}
\begin{tabular}{cc|llll}
\hline
\multicolumn{2}{c|}{Models}                     &  GCN & GIN & GAT & DIG-Mol  \\ \hline
\multicolumn{1}{l|}{Datasets}                  & Tasks &  \multicolumn{4}{c}{5 shots}  \\ \hline
\multicolumn{1}{l|}{\multirow{4}{*}{Tox21}} & SR-HS  & 68.13 & 67.20 & 68.52 & \begin{math}
    \mathbf{70.53}
\end{math} \\
\multicolumn{1}{l|}{} & SR-MMP                 & 69.06  & 70.21& 69.73 & \begin{math}
    \mathbf{73.67}
\end{math} \\
\multicolumn{1}{l|}{} & SR-p53                 & 72.01 & 71.74 & 72.56 & \begin{math}
    \mathbf{75.79}
\end{math}  \\
\multicolumn{1}{l|}{} & Average                & 69.73 & 69.72 & 70.27 & \begin{math}
    \mathbf{73.33}
\end{math} \\ \hline
\multicolumn{1}{l|}{\multirow{7}{*}{SIDER}} & Si-T1 & 65.66 & 66.25 & 67.71 & \begin{math}
    \mathbf{70.35}
\end{math} \\
\multicolumn{1}{l|}{}                  & Si-T2 & 63.62 & 63.10 & 62.94 & \begin{math}
    \mathbf{66.73}
\end{math}  \\
\multicolumn{1}{l|}{}                  & Si-T3 & 61.92 & 63.49 & 63.36 & \begin{math}
    \mathbf{67.46}
\end{math} \\
\multicolumn{1}{l|}{}                  & Si-T4 & 64.85 & 64.09 & 66.62 & \begin{math}
    \mathbf{69.62}
\end{math} \\
\multicolumn{1}{l|}{}                  & Si-T5 & 73.93 & 73.80 & 74.83 & \begin{math}
    \mathbf{77.78}
\end{math} \\
\multicolumn{1}{l|}{}                  & Si-T6 & 62.06 & 63.86 & 64.54 & \begin{math}
    \mathbf{67.17}
\end{math} \\
\multicolumn{1}{l|}{}                  & Average & 64.51 & 65.09 & 66.67 & \begin{math}
    \mathbf{69.85}
\end{math} \\ \hline
\end{tabular}

\raggedright
    *The performances of DIG-Mol for each subtask are marked as bold.
\end{table*}

\subsection{T-SNE visualization}

\begin{figure*}[!htb]
	\centering
	\includegraphics[scale=0.5]{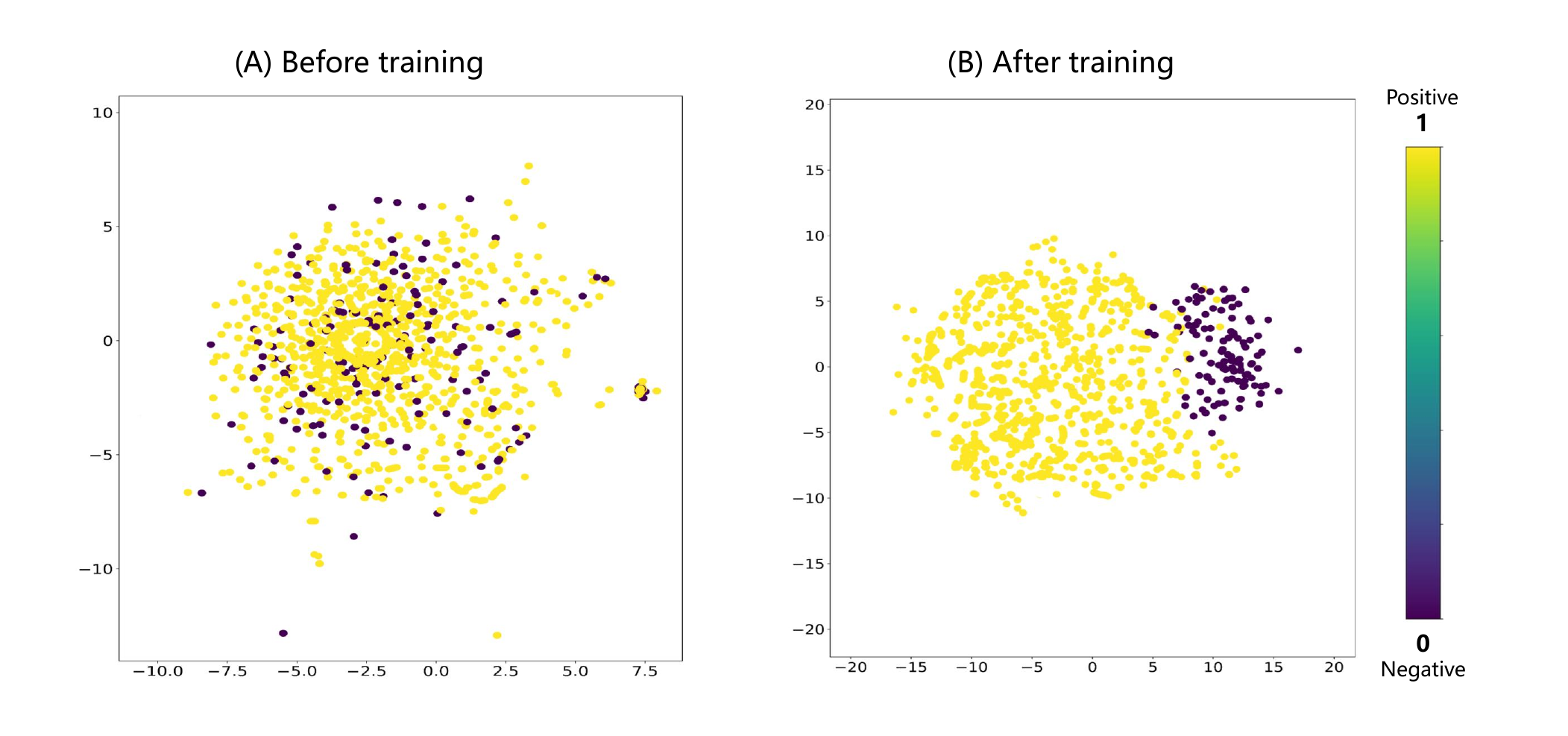}
    \vspace{-5pt}
	\caption{Visualization of spatial localization of molecular representations in the BBBP dataset. The left column displays the initial representations, while the right column showcases pretrained representations }
	\label{fig:Fig 4.}
\end{figure*}

To further substantiate the efficacy and interpretability of DIG-Mol, we employed t-SNE to visualize the molecular representations both pre- and post-training. In Fig \ref{fig:Fig 4.}., we present the 2D embeddings of molecule representations derived from the BBBP dataset. In this visualization, yellow points for molecules with the target property and purple for those without. Initially, the molecular representations are spatially distributed in confusion, with most yellow data points intermingled with purple ones. However, post-training, there is a marked aggregation of same-class molecules and a clear demarcation between the two classes. This change in distribution demonstrates that DIG-Mol effectively captures the physicochemical characteristics of the molecules, resulting in similar latent representations for molecules with similar properties. 

\section{Exploration of chemical interpretation}
In advancing molecular property prediction models for drug discovery, ensuring that these models deliver chemically interpretable results is crucial. This means that the molecular representations they generate during pre-training and fine-tuning should align with established chemical principles and intuitive understanding. Additionally, by leveraging semi-supervised learning in pre-training, the model gains a heightened ability for task-agnostic transfer, which is key in multiple applications. This advantage stems from the model's proficiency in identifying the most influential input features affecting the targeted property in various downstream tasks. Such a capability is invaluable for enhancing our understanding of the underlying factors that determine molecular bioactivity, physiology, and toxicity.

\begin{figure*}[htb]
	\centering
	\includegraphics[scale=0.29]{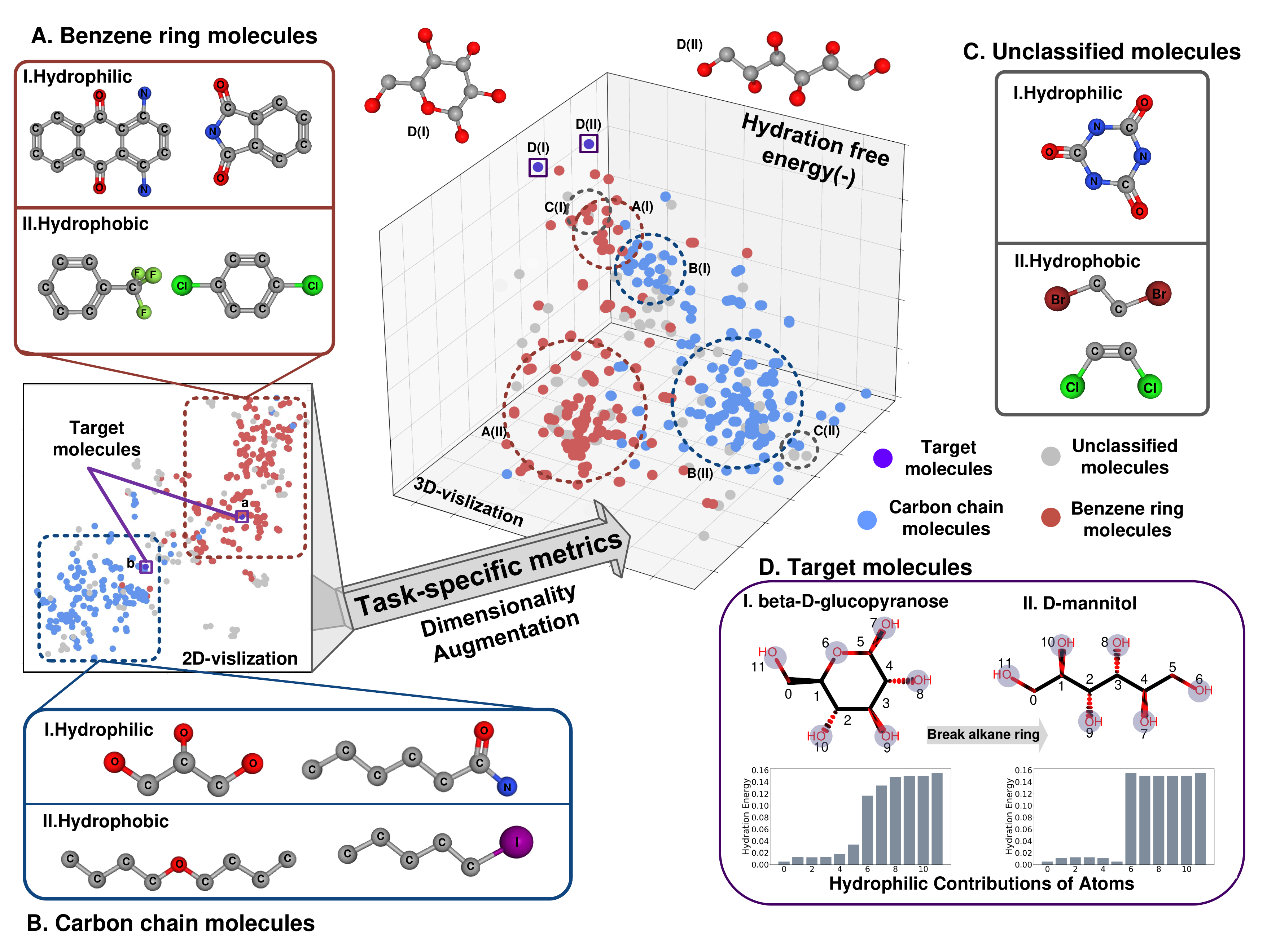}
	\caption{DIG-Mol Visualizations for Exploring Chemical Interpretation using the FreeSolv Dataset. The 2-D visualization depicts molecular representations generated by pre-trained DIG-Mol. The 3-D visualization illustrates molecular representations generated by DIG-Mol during the fine-tuning phase, with the height axis corresponding to the negative values of molecular hydration free energy. Circumambient molecules represent key molecules within hydrophilic and hydrophobic clusters, delineated by dashed circular outlines in the visualization space.}
	\label{fig:Fig 5.}
\end{figure*}

In this context, we employed the T-SNE visualization to confirm the intuitive interpretability of DIG-Mol’s representation outcomes within chemical space. To avoid overly rigid representations, we used the FreeSolv dataset for the regression task. This dataset uses hydration free energy values to shed light on the degree of molecular solvation in aqueous environments 
\cite{frees}. The visualization result is depicted in Figure \ref{fig:Fig 5.}. It is noteworthy that the molecules with lower hydration free energy values tend to be more stable when dissolved in water, indicating a higher solvation level. Furthermore, hydration free energy, as an intrinsic physical property, closely relates to the molecular functional groups’ composition and structural features. This relationship is crucial for enhancing both the understanding and interpretation of the data.

To assess DIG-Mol’s ability to accurately capture crucial features that affect molecular solubility, we employed an innovative and nuanced classification approach to the dataset. This led to the formation of four distinct categories: carbon chain molecules, benzene ring molecules, unclassified molecules (those without benzene rings and carbon chains), and two specific target molecules. The importance of benzene rings and carbon chains as fundamental molecular structures is well-recognized, especially as the functional groups attached to these scaffolds significantly influence solubility in water \cite{frees1}. This is particularly true for the Unclassified molecules. The two selected target molecules, beta-D-glucopyranose (I) and D-mannitol (II), are notable for having the highest solubility in the dataset. Despite having the same elemental compositions and number of carbon atom, they differ in structure:  molecule I has a cyclic form, whereas molecule II has a linear chain configuration.

In the visualization process, we projected the representations generated by the pre-trained DIG-Mol model for the FreeSolv dataset onto a two-dimensional plane. This process highlighted clear differences in the molecular structures of carbon chain molecules and clustering patterns, with each group forming well-defined and separate clusters on the plane. The unclassified molecules, which possess more complex structures, displayed a scattered and random distribution across the plane. Furthermore, the two target molecules, due to their significant structural variances, were represented with a noticeable spatial separation between them, emphasizing the impact of morphological features on their representations.
Subsequently, we presented the molecular representations generated by DIG-Mol during the fine-tuning phase for the FreeSolv dataset, visualized in a three-dimensional space. Unlike conventional methods of increasing dimensionality, this approach retains a two-dimensional projection of molecular representations after reducing dimensionality, while adding a new dimension that represents the negative values of molecular hydration free energy. Essentially, the placement of molecular points along the vertical axis accurately mirrors their respective hydration free energy values. Molecules positioned higher within this three-dimensional arrangement are indicative of greater solubility in water.

In the three-dimensional visualization, the introduction of an additional dimension enhances the chemical correlations and interpretability of the spatial distribution of molecular representation points. Carbon chain and benzene ring molecules both form distinct clusters, with hydrophilic molecules (higher solubility) positioned higher in space, and hydrophobic molecules (lower solubility) positioned lower. Unclassified molecules continue to display a scattered spatial distribution. Notably, the two target molecules, beta-D-glucopyranose and D-mannitol, are positioned at the highest layer, aligning with their outstanding hydration free energy values and solubility. Interestingly, while there is a considerable spatial separation between the two distinct hydrophobic clusters, the two hydrophilic clusters are situated in closer proximity. This pattern is also evident with the two target molecules, which are located near each other. We hypothesize that during the fine-tuning process, DIG-Mol deliberately brought the representations of hydrophilic molecules closer together. This adjustment suggests that DIG-Mol prioritized features impacting molecular solubility over molecular scaffolds for these molecules, thereby reducing the influence of scaffolds on the final representations. In contrast, the representations of hydrophobic molecules, where DIG-Mol could not identify common defining features, remained largely influenced by their structural morphology.

The analysis of representative molecules within hydrophilic and hydrophobic clusters, as depicted in Figure \ref{fig:Fig 5.}., lends further support to the underlying hypothesis. Notably, hydrophilic molecules are characterized by their enriched content of oxygen and nitrogen atoms, indicating a prevalent structural presence of hydroxyl, aldehyde, amino, and amine functional groups. This composition aligns with a well-established theory in the field, suggesting that, unlike the halogen atoms found in hydrophobic molecules, nitrogen and oxygen atoms in these specific functional groups have a stronger propensity to attract water molecules and form hydrogen bonds. Such interactions effectively reduce the intermolecular forces among the molecules when submerged in water, facilitating a more stable and homogeneous dispersion and solubility.

Moreover, the visual representation delineates the contribution of individual atoms to the molecular hydration-free energy for both sets, I. beta-D-glucopyranose and II. D-mannitol, supports the notion that DIG-Mol’s effectiveness is not compromised by variations in morphological structures. These visualizations, serving as an intuitive complement to experimental results, effectively demonstrate DIG-Mol’s proficiency. They highlight DIG-Mol’s ability to capture detailed molecular representations, both globally and locally, during the pre-training phase. This capability lays the groundwork for task-specificity and enhances the model’s transferability. In subsequent tasks, DIG-Mol consistently identifies key features relevant to each task, aligning with established chemical intuition and the interpretation of molecular structures.

\section{Conclusion}
In this study, we proposed a novel self-supervised model called DIG-Mol, designed to address two present obstacles in molecular property prediction and molecular representation learning: the lack of comprehensive and depth of molecular representation learning, along with the model’s limited attention to the specificity of molecular data.

Within DIG-Mol, molecules undergo graphical abstraction, and specialized modules extract crucial representations. Extensive benchmarking experiments underscore DIG-Mol’s outstanding performance, achieving state-of-the-art results in property prediction across a range of tasks. Moreover, few-shot tests highlight its adaptability to downstream tasks. Visualization, parameter optimization, and ablation studies further enhance DIG-Mol’s transparency and interpretability. In summary, DIG-Mol presents a promising and interpretable framework that promises to revolutionize drug discovery by capturing the richness of molecular structure.

\begin{tcolorbox}[sharp corners, colback=white]
    \textbf{• Key points}
\begin{itemize}
    \item DIG-Mol is a novel graph contrastive learning (GCL) framework for molecular property prediction featuring a dual-interaction mechanism and a graph diffusion network. 
\end{itemize}

\begin{itemize}
    \item DIG-Mol leverages Pseudo-siamese networks for self-supervised learning, achieving superior molecular representation through dual contrastive objectives.
\end{itemize}

\begin{itemize}
    \item The architecture of DIG-Mol employs momentum distillation, with the target network guiding the online network to refine representations, enhancing learning efficacy.
\end{itemize}
    
\begin{itemize}
    \item A unique graph augmentation strategy involving unidirectional bond deletion enables the preservation of one-way interactions, crucial for capturing molecular substructures and improving prediction accuracy.
\end{itemize}
\end{tcolorbox}

\section{Funding}
\noindent The National Natural Science Foundation of China (No. 62202388); the National Key Research and Development Program of China (No. 2022YFF1000100); and the Qin Chuangyuan Innovation and Entrepreneurship Talent Project (No. QCYRCXM-2022-230).

\bibliographystyle{unsrt}
\bibliography{reference}

\begin{thebibliography}{10}

\bibitem{intro1}
Linda Martin, Melissa Hutchens, and Conrad Hawkins.
\newblock Trial watch: clinical trial cycle times continue to increase despite industry efforts.
\newblock {\em Nature Reviews Drug Discovery}, 16(3):157--158, 2017.

\bibitem{intro2}
Steven~M Paul, Daniel~S Mytelka, Christopher~T Dunwiddie, Charles~C Persinger, Bernard~H Munos, Stacy~R Lindborg, and Aaron~L Schacht.
\newblock How to improve r\&d productivity: the pharmaceutical industry's grand challenge.
\newblock {\em Nature reviews Drug discovery}, 9(3):203--214, 2010.

\bibitem{intro3}
Olivier~J Wouters, Martin McKee, and Jeroen Luyten.
\newblock Estimated research and development investment needed to bring a new medicine to market, 2009-2018.
\newblock {\em Jama}, 323(9):844--853, 2020.

\bibitem{intro4}
Helen Dowden and Jamie Munro.
\newblock Trends in clinical success rates and therapeutic focus.
\newblock {\em Nat. Rev. Drug Discov}, 18(7):495--496, 2019.

\bibitem{intro6}
Qing Ye, Chang-Yu Hsieh, Ziyi Yang, Yu~Kang, Jiming Chen, Dongsheng Cao, Shibo He, and Tingjun Hou.
\newblock A unified drug--target interaction prediction framework based on knowledge graph and recommendation system.
\newblock {\em Nature communications}, 12(1):6775, 2021.

\bibitem{intro7}
Srivamshi Pittala, William Koehler, Jonathan Deans, Daniel Salinas, Martin Bringmann, Katharina~Sophia Volz, and Berk Kapicioglu.
\newblock Relation-weighted link prediction for disease gene identification.
\newblock {\em arXiv preprint arXiv:2011.05138}, 2020.

\bibitem{intro9}
Jonathan~M Stokes, Kevin Yang, Kyle Swanson, Wengong Jin, Andres Cubillos-Ruiz, Nina~M Donghia, Craig~R MacNair, Shawn French, Lindsey~A Carfrae, Zohar Bloom-Ackermann, et~al.
\newblock A deep learning approach to antibiotic discovery.
\newblock {\em Cell}, 180(4):688--702, 2020.

\bibitem{intro12}
Wen Torng and Russ~B Altman.
\newblock Graph convolutional neural networks for predicting drug-target interactions.
\newblock {\em Journal of chemical information and modeling}, 59(10):4131--4149, 2019.

\bibitem{intro15}
Wei Zheng, Natasha Thorne, and John~C McKew.
\newblock Phenotypic screens as a renewed approach for drug discovery.
\newblock {\em Drug discovery today}, 18(21-22):1067--1073, 2013.

\bibitem{fingerprints1}
Derek~T Ahneman, Jes{\'u}s~G Estrada, Shishi Lin, Spencer~D Dreher, and Abigail~G Doyle.
\newblock Predicting reaction performance in c--n cross-coupling using machine learning.
\newblock {\em Science}, 360(6385):186--190, 2018.

\bibitem{intro18}
J~Ma, RP~Sheridan, A~Liaw, GE~Dahl, and V~Svetnik.
\newblock Deep neural nets as a method for quantitative structure-activity relationships.
\newblock {\em Journal of Chemical Information and Modeling}, 55(2):263--274, 2015.

\bibitem{lfy4}
Fuyi Li, Shuangyu Dong, Andr{\'e} Leier, Meiya Han, Xudong Guo, Jing Xu, Xiaoyu Wang, Shirui Pan, Cangzhi Jia, Yang Zhang, et~al.
\newblock Positive-unlabeled learning in bioinformatics and computational biology: a brief review.
\newblock {\em Briefings in bioinformatics}, 23(1):bbab461, 2022.

\bibitem{nlp1}
Fuxu Wang, Haoyan Wang, Lizhuang Wang, Haoyu Lu, Shizheng Qiu, Tianyi Zang, Xinjun Zhang, and Yang Hu.
\newblock Mhcroberta: pan-specific peptide--mhc class i binding prediction through transfer learning with label-agnostic protein sequences.
\newblock {\em Briefings in Bioinformatics}, 23(3):bbab595, 2022.

\bibitem{intro26}
Jos{\'e} Jim{\'e}nez-Luna, Francesca Grisoni, Nils Weskamp, and Gisbert Schneider.
\newblock Artificial intelligence in drug discovery: recent advances and future perspeves.
\newblock {\em Expert opinion on drug discovery}, 16(9):949--959, 2021.

\bibitem{intro25}
Dejun Jiang, Zhenxing Wu, Chang-Yu Hsieh, Guangyong Chen, Ben Liao, Zhe Wang, Chao Shen, Dongsheng Cao, Jian Wu, and Tingjun Hou.
\newblock Could graph neural networks learn better molecular representation for drug discovery? a comparison study of descriptor-based and graph-based models.
\newblock {\em Journal of cheminformatics}, 13(1):1--23, 2021.

\bibitem{intro35}
Steven Kearnes, Kevin McCloskey, Marc Berndl, Vijay Pande, and Patrick Riley.
\newblock Molecular graph convolutions: moving beyond fingerprints.
\newblock {\em Journal of computer-aided molecular design}, 30:595--608, 2016.

\bibitem{rev1}
Thomas Gaudelet, Ben Day, Arian~R Jamasb, Jyothish Soman, Cristian Regep, Gertrude Liu, Jeremy~BR Hayter, Richard Vickers, Charles Roberts, Jian Tang, et~al.
\newblock Utilizing graph machine learning within drug discovery and development.
\newblock {\em Briefings in bioinformatics}, 22(6):bbab159, 2021.

\bibitem{gcn2}
Mengying Sun, Sendong Zhao, Coryandar Gilvary, Olivier Elemento, Jiayu Zhou, and Fei Wang.
\newblock Graph convolutional networks for computational drug development and discovery.
\newblock {\em Briefings in bioinformatics}, 21(3):919--935, 2020.

\bibitem{intro16}
Nathan Brown, Marco Fiscato, Marwin~HS Segler, and Alain~C Vaucher.
\newblock Guacamol: benchmarking models for de novo molecular design.
\newblock {\em Journal of chemical information and modeling}, 59(3):1096--1108, 2019.

\bibitem{gcl1}
Yuning You, Tianlong Chen, Yongduo Sui, Ting Chen, Zhangyang Wang, and Yang Shen.
\newblock Graph contrastive learning with augmentations.
\newblock {\em Advances in neural information processing systems}, 33:5812--5823, 2020.

\bibitem{kg1}
Yin Fang, Qiang Zhang, Haihong Yang, Xiang Zhuang, Shumin Deng, Wen Zhang, Ming Qin, Zhuo Chen, Xiaohui Fan, and Huajun Chen.
\newblock Molecular contrastive learning with chemical element knowledge graph.
\newblock In {\em Proceedings of the AAAI Conference on Artificial Intelligence}, volume~36, pages 3968--3976, 2022.

\bibitem{gat2}
Hui Liu, Yibiao Huang, Xuejun Liu, and Lei Deng.
\newblock Attention-wise masked graph contrastive learning for predicting molecular property.
\newblock {\em Briefings in bioinformatics}, 23(5):bbac303, 2022.

\bibitem{cas}
Zixi Zheng, Yanyan Tan, Hong Wang, Shengpeng Yu, Tianyu Liu, and Cheng Liang.
\newblock Casangcl: pre-training and fine-tuning model based on cascaded attention network and graph contrastive learning for molecular property prediction.
\newblock {\em Briefings in Bioinformatics}, 24(1):bbac566, 2023.

\bibitem{nat1}
Yuyang Wang, Jianren Wang, Zhonglin Cao, and Amir Barati~Farimani.
\newblock Molecular contrastive learning of representations via graph neural networks.
\newblock {\em Nature Machine Intelligence}, 4(3):279--287, 2022.

\bibitem{aug1}
Omar Mahmood, Elman Mansimov, Richard Bonneau, and Kyunghyun Cho.
\newblock Masked graph modeling for molecule generation.
\newblock {\em Nature communications}, 12(1):3156, 2021.

\bibitem{pubchem}
Sunghwan Kim, Jie Chen, Tiejun Cheng, Asta Gindulyte, Jia He, Siqian He, Qingliang Li, Benjamin~A Shoemaker, Paul~A Thiessen, Bo~Yu, et~al.
\newblock Pubchem 2019 update: improved access to chemical data.
\newblock {\em Nucleic acids research}, 47(D1):D1102--D1109, 2019.

\bibitem{gcl}
Ming Jin, Yizhen Zheng, Yuan-Fang Li, Chen Gong, Chuan Zhou, and Shirui Pan.
\newblock Multi-scale contrastive siamese networks for self-supervised graph representation learning.
\newblock {\em arXiv preprint arXiv:2105.05682}, 2021.

\bibitem{lfy2}
Yan Zhu, Fuyi Li, Xudong Guo, Xiaoyu Wang, Lachlan~JM Coin, Geoffrey~I Webb, Jiangning Song, and Cangzhi Jia.
\newblock Timer is a siamese neural network-based framework for identifying both general and species-specific bacterial promoters.
\newblock {\em Briefings in Bioinformatics}, page bbad209, 2023.

\bibitem{diffusion}
Zonghan Wu, Shirui Pan, Guodong Long, Jing Jiang, and Chengqi Zhang.
\newblock Graph wavenet for deep spatial-temporal graph modeling.
\newblock {\em arXiv preprint arXiv:1906.00121}, 2019.

\bibitem{dcrnn}
Yaguang Li, Rose Yu, Cyrus Shahabi, and Yan Liu.
\newblock Diffusion convolutional recurrent neural network: Data-driven traffic forecasting.
\newblock {\em arXiv preprint arXiv:1707.01926}, 2017.

\bibitem{conloss}
Prannay Khosla, Piotr Teterwak, Chen Wang, Aaron Sarna, Yonglong Tian, Phillip Isola, Aaron Maschinot, Ce~Liu, and Dilip Krishnan.
\newblock Supervised contrastive learning.
\newblock {\em Advances in neural information processing systems}, 33:18661--18673, 2020.

\bibitem{moleculenet}
Zhenqin Wu, Bharath Ramsundar, Evan~N Feinberg, Joseph Gomes, Caleb Geniesse, Aneesh~S Pappu, Karl Leswing, and Vijay Pande.
\newblock Moleculenet: a benchmark for molecular machine learning.
\newblock {\em Chemical science}, 9(2):513--530, 2018.

\bibitem{schnet}
Kristof~T Sch{\"u}tt, Huziel~E Sauceda, P-J Kindermans, Alexandre Tkatchenko, and K-R M{\"u}ller.
\newblock Schnet--a deep learning architecture for molecules and materials.
\newblock {\em The Journal of Chemical Physics}, 148(24), 2018.

\bibitem{mgcn}
Chengqiang Lu, Qi~Liu, Chao Wang, Zhenya Huang, Peize Lin, and Lixin He.
\newblock Molecular property prediction: A multilevel quantum interactions modeling perspective.
\newblock In {\em Proceedings of the AAAI conference on artificial intelligence}, volume~33, pages 1052--1060, 2019.

\bibitem{dmpnn}
Kevin Yang, Kyle Swanson, Wengong Jin, Connor Coley, Philipp Eiden, Hua Gao, Angel Guzman-Perez, Timothy Hopper, Brian Kelley, Miriam Mathea, et~al.
\newblock Analyzing learned molecular representations for property prediction.
\newblock {\em Journal of chemical information and modeling}, 59(8):3370--3388, 2019.

\bibitem{ngarm}
Shengchao Liu, Mehmet~F Demirel, and Yingyu Liang.
\newblock N-gram graph: Simple unsupervised representation for graphs, with applications to molecules.
\newblock {\em Advances in neural information processing systems}, 32, 2019.

\bibitem{hu}
Weihua Hu, Bowen Liu, Joseph Gomes, Marinka Zitnik, Percy Liang, Vijay Pande, and Jure Leskovec.
\newblock Strategies for pre-training graph neural networks.
\newblock {\em arXiv preprint arXiv:1905.12265}, 2019.

\bibitem{gcn1}
Si~Zhang, Hanghang Tong, Jiejun Xu, and Ross Maciejewski.
\newblock Graph convolutional networks: a comprehensive review.
\newblock {\em Computational Social Networks}, 6(1):1--23, 2019.

\bibitem{ginfew}
Zhichun Guo, Chuxu Zhang, Wenhao Yu, John Herr, Olaf Wiest, Meng Jiang, and Nitesh~V Chawla.
\newblock Few-shot graph learning for molecular property prediction.
\newblock In {\em Proceedings of the web conference 2021}, pages 2559--2567, 2021.

\bibitem{frees}
David~L Mobley and J~Peter Guthrie.
\newblock Freesolv: a database of experimental and calculated hydration free energies, with input files.
\newblock {\em Journal of computer-aided molecular design}, 28:711--720, 2014.

\bibitem{frees1}
Gerhard Hummer, Lawrence~R Pratt, and Angel~E Garcia.
\newblock Free energy of ionic hydration.
\newblock {\em The Journal of Physical Chemistry}, 100(4):1206--1215, 1996.

\bibitem{rf}
Yanjun Qi.
\newblock Random forest for bioinformatics.
\newblock {\em Ensemble machine learning: Methods and applications}, pages 307--323, 2012.

\bibitem{svm}
Marti~A. Hearst, Susan~T Dumais, Edgar Osuna, John Platt, and Bernhard Scholkopf.
\newblock Support vector machines.
\newblock {\em IEEE Intelligent Systems and their applications}, 13(4):18--28, 1998.

\end{thebibliography}

\newpage
\newpage

\clearpage
\setcounter{page}{1}
\section{Supplementary Information}

\subsection{Appendix \textup{I}}

In the GNN, the node embeddings \begin{math}
\mathcal{H}\end{math}, GNN encoder \begin{math}g\end{math} utilizes a neighborhood aggregation operation, which updates the node representation iteratively. Such neighborhood aggregation for the feature of one node on the \begin{math}
    k-1
\end{math} layer of a GNN encoder can be formulated as:
\begin{align}
\label{eq:Aeq1}
\mathcal{H}_{i}^{(k)}=\sigma(\mathcal{H}_{i}^{(k-1)},\;f^{(k)}(\{\mathcal{H}_{j}^{(k-1)}:\forall j\in \mathcal{N}(i) \}))
\end{align}

In our work, the graphs we need to process are molecule graphs, to utilize the GNN encoder, molecule structures are transformed into heterogeneous graph by RDkit. A molecule graph \begin{math}\mathcal{G}\end{math} is defined as \begin{math}\mathcal{G =(\Large{\upsilon},\Large{\varepsilon}) }\end{math}, where \begin{math}\mathcal{\upsilon}\end{math} and \begin{math}\mathcal{\varepsilon}\end{math} are atoms (nodes) and chemical bonds (edges), respectively. In the iterative process, the GNN encoder follows an explicit message-passing paradigm, the representation of each atom is updated by aggregating their neighborhood-- other atoms that are linked with them through chemical bonds. The updated operations perform iteratively and further spread to all the atoms. After several iterations of aggregation operations, the high-order representation information of atoms can be captured. The aggregation update rule for an atom feature on the k-th iterations is represented as
\begin{align}
\label{eq:Aeq2}
a _{i}^{(k)}=AGGREGATE^{(k)}\left ( \left\{\mathcal{H}_{j}^{(k-1)}:\forall j\in \mathcal{N}(i)\right\}\right)
\end{align}

\begin{align}\label{eq}
\mathcal{H}_{i}^{(k)}=COMBINE^{(k)}\left ( \mathcal{H}_{i}^{(k-1)},a _{i}^{(k)}  \right )\end{align}
where \begin{math}
\mathcal{H}_{i}^{(k)}    
\end{math} is the feature of node \begin{math}
\mathbf{i} 
\end{math} at the k-th iterations and \begin{math}
\mathcal{H}_{i}^{(0)}    
\end{math} is initialized by node feature\begin{math}
\mathbf{x} _{i}    
\end{math}. And AGGREGATE(·) and COMBINE(·) are two essential operators that are used to extract and aggregate adjacent information of node \begin{math}
i
\end{math}.
To further extract a graph-level representation \begin{math}
\mathcal{H} _{\mathcal{G}}    
\end{math}, a readout operation is employed to integrate all the atom's features and obtain graph representation. 

\begin{align}\label{eq2}
\mathcal{H}_{\mathcal{G}}=READOUT\left ( \left\{\mathcal{H}_{i}^{(k)}: i\in \mathcal{G}\right\}\right) 
\end{align}

\subsection{Appendix \textup{II}}
Table \ref{tab:Atab 1.}. contains all notations used in this paper, along with their derivations and explanations. The notations are categorized into three sections: the overall framework, diffusion graph network, contrastive loss calculation, and appendix. Within each section, notations are arranged in the order of their appearance in the paper. Notations that are mentioned and deemed important but not specifically tied to a particular equation are also included in the table.
\begin{table}[ht]
    \centering
    \caption{Detailed information of all notations used in this paper.}
    \label{tab:Atab 1.}
        \begin{tabular}{c|c|c}
        \toprule[1pt]
            Notation & Derivation & Explanation \\ \hline
            \begin{math}
                \mathcal{D}
            \end{math} & Eq.\ref{eq:Eq1} & Set of molecules SMILES strings. \\
            \begin{math}
                \mathcal{S}
            \end{math} & Eq.\ref{eq:Eq1} & Molecules SMILES string. \\
            \begin{math}
                \mathcal{D}_{\mathcal{G}}
            \end{math} & Eq.\ref{eq:Eq1} & Set of molecular graphs. \\
            \begin{math}
                \mathcal{G}
            \end{math} & Eq.\ref{eq:Eq1} & Molecular graph. \\ 
            \begin{math}
                n
            \end{math} & Eq.\ref{eq:Eq1} & Number of molecules in set. \\
            \begin{math}
                \mathcal{Y}
            \end{math} & Eq.\ref{eq:Eq2} & Labeled information of molecules. \\
            \begin{math}
                (\mathcal{X}, \mathcal{A})
            \end{math} & - & \makecell{Node feature matrix and \\ adjaency matrix.} \\
            \begin{math}
                (\upsilon, \varepsilon)
            \end{math} & - & Nodes and edges in graph. \\
            \begin{math}
                \mathcal{G}^{1},\mathcal{G}^{2}
            \end{math} & Eq.\ref{eq:Eq2} & Two augmented graphs. \\
            \begin{math}
                g_{\theta},g_{\xi}
            \end{math} & - & \makecell{Graph encoder of online network \\ and target network.} \\
            \begin{math}
                p_{\theta},p_{\xi}
            \end{math} & - & \makecell{Projection head encoder of online \\ network and target network.} \\
            \begin{math}
                t
            \end{math} & Eq.\ref{eq:Eq6} & The number of training epochs.  \\
            \begin{math}
                m
            \end{math} & Eq.\ref{eq:Eq7} & The rate of dynamical updating. \\
            \begin{math}
                \mathcal{H}
            \end{math} & Eq.\ref{eq:Eq7} & Graph or node embeddings. \\
            \begin{math}
                \mathcal{Z}_{\theta}^{1},\mathcal{Z}_{\theta}^{2}
            \end{math} & - & \makecell{The final projection vectors \\ of online network.} \\
            \begin{math}
                \mathcal{Z}_{\xi}^{1},\mathcal{Z}_{\xi}^{2}
            \end{math} & - & \makecell{The final projection vectors \\ of otarget network.} \\
            \begin{math}
             N,M,P
            \end{math} & - & Matrix dimensions. \\
            \begin{math}
               \widetilde{\mathcal{A}}  
            \end{math} & Eq.\ref{eq:Eq8} & \makecell{The normalized adjacent matrix \\ with self loops.}\\
            \begin{math}
                \mathcal{W} 
            \end{math} & Eq.\ref{eq:Eq8} & \makecell{The learnable matrix in \\ convolution operation. } \\ \hline
            \begin{math}
                \mathcal{P}
            \end{math} & Eq.\ref{eq:Eq9} & \makecell{The transition matrix in \\diffusion process.} \\
            \begin{math}
                \eta
            \end{math} & Eq.\ref{eq:Eq9} & \makecell{The restart probability of random \\message passing.} \\
            \begin{math}
                W
            \end{math} & Eq.\ref{eq:Eq9} & \makecell{The learnable matrix in\\ diffusion operation.} \\

            \begin{math}
                K
            \end{math} & Eq.\ref{eq:Eq10} & \makecell{The number of diffusion steps.} \\
            \begin{math}
                f_{\phi },\phi
            \end{math} & Eq.\ref{eq:Eq7} & \makecell{The graph filter and its parameter \\ in diffusion process.} \\
            \begin{math}
                \epsilon
            \end{math} & Eq.\ref{eq:Eq12} & Constant coefficient. \\ \hline
            \begin{math}
                \tau 
            \end{math} & Eq.\ref{eq:Eq14} & Temperature parameter. \\
            \begin{math}
                B 
            \end{math} & Eq.\ref{eq:Eq14} & Batch size. \\
            \begin{math}
                \mathcal{L}_{GI}
            \end{math} & Eq.\ref{eq:Eq14} & Graph-interaction contrastive loss.  \\
            \begin{math}
                \mathcal{L}_{EI}
            \end{math} & - & Encoder-interaction contrastive loss. \\
            \begin{math}
              \mathcal{L}_{MI}
            \end{math} & - & Multi-interaction contrastive loss. \\
             \begin{math}
                \mathcal{L}_{jiont}
            \end{math}& Eq.\ref{eq:Eq17} & Joint contrastive loss. \\
            \begin{math}
                \alpha,\beta,\gamma 
            \end{math} & Eq.\ref{eq:Eq17} & Independent constrative loss. \\ \hline
            \begin{math}
                \mathcal{N}(.)
            \end{math} & Eq.\ref{eq:Aeq1} & Node neighbours in graph.\\
            \begin{math}
                \sigma(.)
            \end{math} & Eq.\ref{eq:Aeq1} &  Activation function. \\ 
            \begin{math}
                a
            \end{math} & Eq.\ref{eq:Aeq2} & Aggregated node feature. \\
            \hline
        \end{tabular}
\end{table}

\subsection{Appendix \textup{III}}
Algorithm \ref{alg:Alg 1.} provides a comprehensive account of the steps involved in the pretraining stage of DIG-Mol, encompassing the entire process from molecules SMILES input to contrasitive loss calculation and the completion of training.

\begin{algorithm}[ht]
\caption{Pseudocode of DIG-Mol pretraining.}
\label{alg:Alg 1.}
\hspace*{0.01in} {\bf Input:} 
A set of molecule SMILES strings \begin{math}
    \mathcal{D}=\left \{ \mathcal{S}_{i}\right \} _{i=1} ^{\left | \mathcal{D} \right | }\end{math}\\
\hspace*{0.01in} {\bf } Molecular augmentation operation: \begin{math}
    \mathcal{A}\end{math}\\
\hspace*{0.01in} {\bf } 
Online encoder and projector with parameter \begin{math}
    \theta: g_{\theta}, p_{\theta}
\end{math}\\
\hspace*{0.01in} {\bf }
Target encoder and projector with parameter 
\begin{math}
    \xi: g_{\xi}, p_{\xi}
\end{math}\\
\hspace*{0.01in} {\bf Output:} 
Learnable parameter of online encoder \begin{math}
    g_{\theta}\end{math}
\begin{algorithmic}[1]
\State Transfer molecule SMILES strings to molecular graphs.
\State \begin{math}
    \mathcal{D}=\left \{ \mathcal{S}_{i}\right \} _{i=1} ^{\left | \mathcal{D} \right | }\to \mathcal{D}_{\mathcal{G}}=\left \{ \mathcal{G}_{i}\right \} _{i=1} ^{\left | \mathcal{D} \right | }
\end{math}
\State \begin{math}
    \mathcal{G}_{i}=\left \{ \mathcal{X}_{1},\mathcal{X}_{2}\dots \mathcal{X}_{N} \right \}\in \mathcal{G}
\end{math}
\State Randomly initialize the parameters \begin{math}
    \theta: g_{\theta}, p_{\theta}\end{math}
\State Deep copy \begin{math} g_{\theta} \end{math} and \begin{math} p_{\theta} \end{math} to \begin{math} g_{\xi} \end{math} and \begin{math} p_{\xi} \end{math}
\State ***Training phase***
\For{\begin{math}
    \mathcal{G}_{i}\in \mathcal{G}
\end{math}}
\State Molecular graphs augmentation \begin{math}
    \mathcal{G}_{i}\overset{\mathcal{A}}{\to } \left \{\mathcal{G}_{i}^{1}, \mathcal{G}_{i}^{2}\right \}
\end{math}
\State Graph latent embeddings
\State\begin{math}
\mathcal{H}_{\theta}^1= g_{\theta}(\mathcal{G}_{i}^{1}), \mathcal{H}_{\theta}^2= g_{\theta}(\mathcal{G}_{i}^{2})
\end{math}
\State \begin{math}
\mathcal{H}_{\xi}^1= g_{\xi}(\mathcal{G}_{i}^{1}), \mathcal{H}_{\xi}^2= g_{\xi}(\mathcal{G}_{i}^{2})
\end{math}
\State Projection head conversion 
\State \begin{math}
\mathcal{Z}_{\theta}^1= p_{\theta}(\mathcal{H}_{\theta}^1)=p_{\theta}(g_{\theta}(\mathcal{G}_{i}^{1})), \mathcal{Z}_{\theta}^2= p_{\theta}(\mathcal{H}_{\theta}^2)=p_{\theta}(g_{\theta}(\mathcal{G}_{i}^{2}))
\end{math}
\State \begin{math}
\mathcal{Z}_{\xi}^1= p_{\xi}(\mathcal{H}_{\xi}^1)=p_{\xi}(g_{\xi}(\mathcal{G}_{i}^{1})), \mathcal{Z}_{\xi}^2= p_{\xi}(\mathcal{H}_{\xi}^2)=p_{\xi}(g_{\xi}(\mathcal{G}_{i}^{2}))
\end{math}
\State Calculate Graph-Interaction loss \begin{math}
    \mathcal{L} _{\small {GI}} \end{math} by Eq.11
\State Calculate Encoder-Interaction loss \begin{math}
    \mathcal{L} _{\small {EI}} \end{math} by Eq.12
\State Calculate Multi-Interaction loss \begin{math}
    \mathcal{L} _{\small {MI}} \end{math} by Eq.13
\State Joint loss \begin{math}
    \mathcal{L} _{\small {Joint}} = \alpha\mathcal{L} _{\small {GI}} + \beta \mathcal{L} _{\small {EI}} + \gamma \mathcal{L} _{\small {MI}}
\end{math}
\EndFor\\
{\bf Begin:} 
Backward Propagation

\State \qquad Compute adapted parameter \begin{math}
    \theta\end{math} with gradient descent
\State \qquad Update parameter \begin{math}
    \theta\end{math} by Adam optimizer
\State \qquad Update the parameters \begin{math}
    \xi \end{math} of target network by Eq.8\\
{\bf End}
\State \Return online encoder parameter \begin{math}
    g_{\theta}\end{math} 
\end{algorithmic}
\end{algorithm}

\subsection{Appendix \textup{IV}}
Table \ref{tab:Atab 2.}. provides comprehensive details concerning the seven classification task datasets and six regression task datasets, encompassing task types, categories, partitioning methods, and so on."

\subsubsection{Classification tasks}

\noindent\begin{math}
 \mathbf{BBBP:}     
\end{math} 
Blood–brain barrier penetration (membrane permeability). This dataset is specifically designed for modeling and predicting barrier permeability, evaluating the ability of over 2000 compounds to penetrate the blood-brain barrier.

\noindent\begin{math}
 \mathbf{BACE:}     
\end{math}
This dataset is a collection of quantitative and qualitative binding results for a set of inhibitors of human \begin{math}
\beta\end{math}-secretase 1, including of 1522 compounds with 2D structures, IC50 and binary qualitative classification results.

\noindent\begin{math}
 \mathbf{SIDER:}     
\end{math}
Side Effect Resource. This dataset is a database of market drugs and adverse drug reactions (ADRs). The SIDER dataset classifies drug side effects into 27 organ categories based on the MedDRA classification, and it includes 1427 approved drugs with their SMILES and binary labels.

\noindent\begin{math}
 \mathbf{ClinTox:}     
\end{math}
The dataset compares FDA-approved drugs with non-approved drugs due to toxicity, containing 1478 drug compounds with known structure and two classification tasks: \begin{math}
    (a)
\end{math} clinical drug toxicity information and \begin{math}
    (b)
\end{math} FDA approval status.

\noindent\begin{math}
 \mathbf{Tox21:}     
\end{math}
Toxicology in the 21st Century. This dataset is a public database for measuring compound toxicity. This dataset includes toxicity measurements of 7831 synthesized substances and structure information for 12 different binary classification tasks.

\noindent\begin{math}
 \mathbf{MUV:}     
\end{math}
Maximum Unbiased Validation. This dataset is selected from the PubChem BioAssay benchmark dataset using refined nearest-neighbor analysis. The MUV dataset includes 17 challenge tasks with about 90,000 compounds, which it used to validate virtual screening techniques.

\noindent\begin{math}
 \mathbf{HIV:}     
\end{math}
Human Immunodeficiency Virus. This dataset originates from the Drug Therapeutics Program (DTP) AIDS Antiviral Screen, which tests over 40000 compounds for their ability to inhibit HIV replication. Screening results are categorized as confirmed inactive (CI), confirmed active (CA), or confirmed moderately active (CM).

\subsubsection{Regression tasks}

\begin{table*}[!ht]
    \setlength{\tabcolsep}{4.5mm}
    \centering
    \caption{Detailed information of 7 classification task benchmarks and 6 regression task benchmarks.}
    \label{tab:Atab 2.}
    \begin{tabular}{llllllll}
        \toprule
        Task type & Datasets & Category & \#Molecules & \#Tasks & Metrics & Split \\
        \hline
        \multirow{7}{*}{Classification} & BBBP & Physiology & 2039 & 1 & ROC-AUC & Scaffold \\
        & BACE & Biophysics & 1513 & 1 & ROC-AUC & Scaffold \\
        & SIDER & Physiology & 1427 & 27 & ROC-AUC & Scaffold \\
        & ClinTox & Physiology & 1478 & 2 & ROC-AUC & Scaffold \\
        & HIV & Biophysics & 41127 & 1 & ROC-AUC & Scaffold \\
        & Tox21 & Physiology & 7831 & 12 & ROC-AUC & Scaffold \\
        & MUV & Biophysics & 93087 & 17 & ROC-AUC & Scaffold \\
        \midrule
        \multirow{6}{*}{Regression} & FreSolv & Physical chemistry & 642 & 1 & RMSE & Scaffold \\
        & ESOL & Physical chemistry & 1128 & 1 & RMSE & Scaffold \\
        & Lipo & Physical chemistry & 4200 & 1 & RMSE & Scaffold \\
        & QM7 & Quantum mechanics & 6830 & 1 & MAE & Scaffold \\
        & QM8 & Quantum mechanics & 21786 & 12 & MAE & Scaffold \\
        & QM9 & Quantum mechanics & 130829 & 8 & MAE & Random \\
        \bottomrule
    \end{tabular}
\end{table*}

\noindent\begin{math}
 \mathbf{FreeSolv:}     
\end{math}
Free Solvation Database. This dataset contains both calculated and experimental hydration free energy values for 642 small molecules in water, with the former being obtained through alchemical free energy calculations using molecular dynamics simulations.

\noindent\begin{math}
 \mathbf{ESOL:}     
\end{math}
Estimated SOLubility. This dataset contains SMILES sequence and water solubility (log solubility in moles per liter) data of 1128 drug molecules. 

\noindent\begin{math}
 \mathbf{Lipo:}     
\end{math}
Lipophilicity. This dataset provides the experimental results of the octanol/water partition coefficient (logD at pH7.4) of more than 4200 compounds from the ChEMBL database.

\noindent\begin{math}
 \mathbf{QM7:}     
\end{math}
Quantum Mechanics 7. This dataset is a subset of GDB-13 (a database of nearly 1 billion stable and synthetically accessible organic molecules) composed of all molecules of up to 23 atoms (including 7 heavy atoms C, N, O, and S), totaling 6830 molecules. It provides the Coulomb matrix representation of these molecules and their atomization energies computed similarly to the FHI-AIMS implementation of the Perdew-Burke-Ernzerhof hybrid functional (PBE0).

\noindent\begin{math}
 \mathbf{QM8:}     
\end{math}
Quantum Mechanics 8. This dataset originates from a recent study focused on the modeling of quantum mechanical calculations pertaining to electronic spectra and excited state energy of small molecules, encompassing a comprehensive collection of four distinct properties related to excited states across 22 thousand samples.

\noindent\begin{math}
 \mathbf{QM9:}     
\end{math}
Quantum Mechanics 9. This dataset is the subset of all 133,885 species with up to nine heavy atoms (C, O, N and F) out of the GDB-17 chemical space of 166 billion organic molecules. The QM9 dataset contains computed geometric, energetic, electronic, and thermodynamic properties for 134k stable small organic molecules, including geometries minimal in energy, corresponding harmonic frequencies, dipole moments, polarizabilities, along with energies, enthalpies, and free energies of atomization.

\subsection{Appendix \textup{V}}
Table \ref{tab:Atab 3.}. provides an overview of all the hyper-parameters used in fine-tuning our model in this specific case.

\begin{table*}[ht]
    \centering
    \caption{Detailed information of hyper-parameters used in the fine-tuning stage.}
    \label{tab:Atab 3.}
    \setlength{\tabcolsep}{5.5mm}
    \begin{tabular}{lll}
    \toprule[1pt]
    Name & Description & Range \\ \hline
    epochs & Fine-tune epochs & \begin{math}
        \left \{{50, 100, 300, 500} \right \}
    \end{math}  \\
    batch\_size & The input batch\_size & \begin{math}
        \left \{{16, 32, 64, 128, 256} \right \}
    \end{math}  \\ 
    emb\_dim & The embedding dimension & \begin{math}
        \left \{{128, 256, 300, 512} \right \}
    \end{math}  \\ 
    lr & The learning rate & \begin{math}
        \left \{{0.0001-0.01} \right \}
    \end{math}  \\
    num\_layer & Number of GNN layers & \begin{math}
        \left \{3, 4, 5, 6, 7\right \}
    \end{math}  \\
    dropout & Dropout rate & \begin{math}
        \left \{{0, 0.1, 0.3, 0.5, 0.7} \right \}
    \end{math}  \\
    temperature & The temperature & \begin{math}
        \left \{{0.05-0.6} \right \}
    \end{math}  \\
    \hline
        \end{tabular}
\end{table*}

\begin{figure*}[h!]
	\centering
	\includegraphics[scale=0.240]{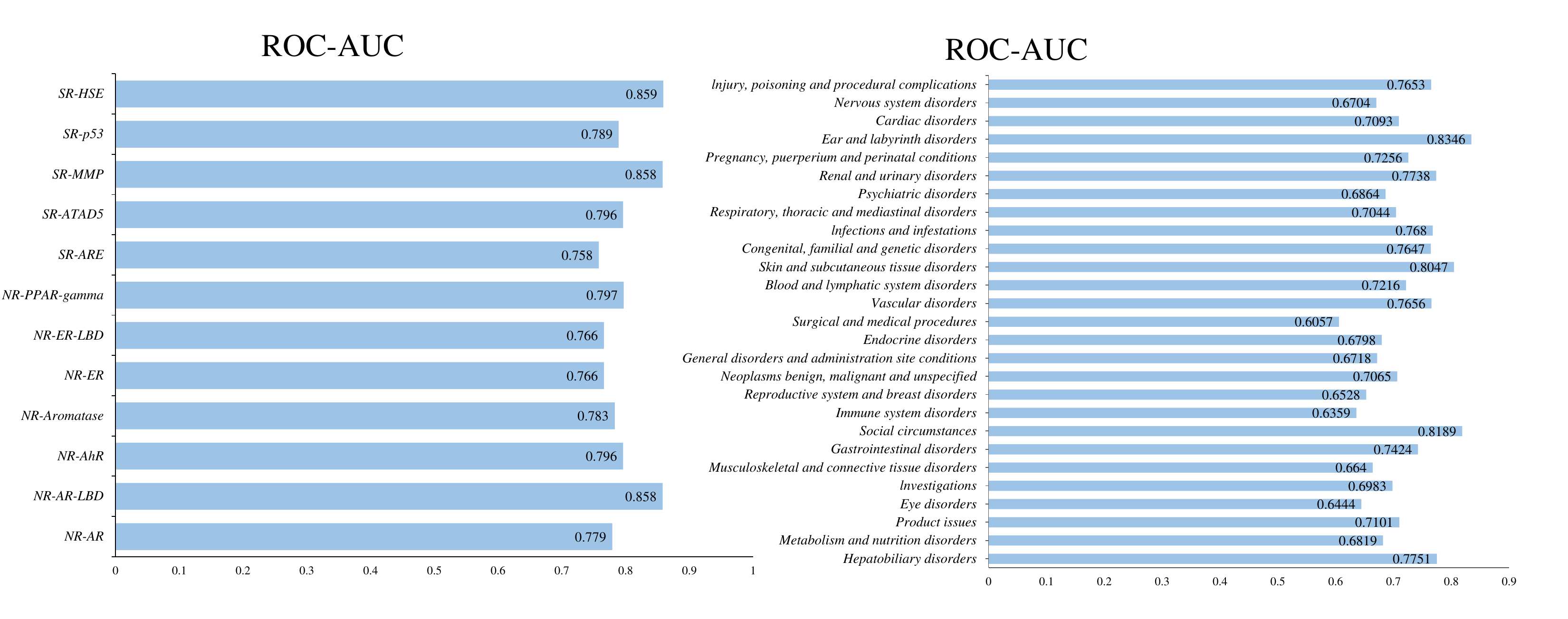}
 \vspace{-5pt}
	\caption{Performance of DIG-Mol on each subtasks within SIDER and Tox21 datasets. }
	\label{fig:Fig 8.}
\end{figure*}
\vspace{-5pt}

\subsection{Appendix \textup{VI}}

\subsubsection{Supervised learning models}

\noindent\begin{math}
 \mathbf{RF}     
\end{math} \cite{rf}:
Random Forest. It is a prediction method based on deep neural networks and uses molecular fingerprints as input. It consists of multiple decision trees, and their prediction results are averaged to obtain the final result.

\noindent\begin{math}
 \mathbf{SVM}     
\end{math} \cite{svm}: 
Support Vector Machine. It is a data-centric approach designed to identify a classification hyperplane for the purpose of efficiently segregating data points belonging to distinct classes.

\noindent\begin{math}
 \mathbf{GCN}     
\end{math} \cite{gcn1}:
Graph Convolutional Network. It is an adapted variant of the conventional Graph Neural Network (GNN) encoder. It seamlessly incorporates convolutional layers into the message-passing framework to effectively extract both the topological attributes and spatial structure of the underlying graph.

\noindent\begin{math}
 \mathbf{GIN}     
\end{math} \cite{ginfew}: 
Graph Isomorphism Network. It is also a variant of the GNN encoder. Similar to the Weisfeiler-Lehman(WL) graph isomorphism test, GIN also aggregates neighborhood node features to recursively update node feature vectors to distinguish different graphs.

\noindent\begin{math}
 \mathbf{SchNet}     
\end{math} \cite{schnet}: 
SchNet is an extension of the DTNN (Deep Tensor Neural Network) framework, uniquely equipped to preserve atomic system rotation and translation invariance, along with atomic permutation invariance during the learning process. It harnesses the power of deep neural networks to forecast molecular properties and interactions by taking into account the spatial arrangement of atoms within molecules.

\noindent\begin{math}
 \mathbf{MGCN}    
\end{math} \cite{mgcn}: Multilevel GCN. The method is a hierarchical graph neural network that directly extracts features from conformation and spatial information, enhancing multilevel interactions. Consequently, it strengthens message passing between nodes and edges in the internal structure of a molecule.

\noindent\begin{math}
 \mathbf{D-MPNN}     
\end{math} \cite{dmpnn}:
Directed Message Passing Neural Network. D-MPNN utilizes directed edge messages to prevent totters and unnecessary loops in message passing, making it more similar to belief propagation in probabilistic graphical models compared to atom-based approaches.

\subsubsection{Self-supervised learning models}

\noindent\begin{math}
 \mathbf{N-Gram}     
\end{math} \cite{ngarm}:
This method incorporates atom embeddings by utilizing their attribute structure and enumerates multiple short walks in the graph. Subsequently, it forms embeddings for each of these walks by aggregating the vertex embeddings. The ultimate representation is derived from the collective embeddings of all the walks.

\noindent\begin{math}
 \mathbf{Hu\;et.al}     
\end{math} \cite{hu}:
This approach serves as a pre-training strategy for GNNs. Its primary focus lies in the pre-training of a highly expressive GNN, incorporating both node and graph-level information. By doing so, it empowers the GNN to acquire meaningful local and global representations simultaneously. This approach promotes the capture of domain-specific semantics effectively.

\noindent\begin{math}
 \mathbf{MolCLR}     
\end{math} \cite{nat1}: Molecular Contrastive Learning of Representations. It is a method that utilizes GNN encoders to acquire differentiable representations during the pre-training phase. It achieves this by constructing molecular graphs and employing a contrastive learning approach, which involves using augmented molecular graphs generated through three distinct augmentation strategies.

\subsection{Appendix \textup{VII}}
Figure \ref{fig:Fig 8.}. illustrates the performance of DIG-Mol on each subtask within SIDER and Tox21 datasets.

\section{Supplementary experiments}

\subsection{Parameter study}

\begin{figure*}[htb]
	\centering
	\includegraphics[scale=0.29]{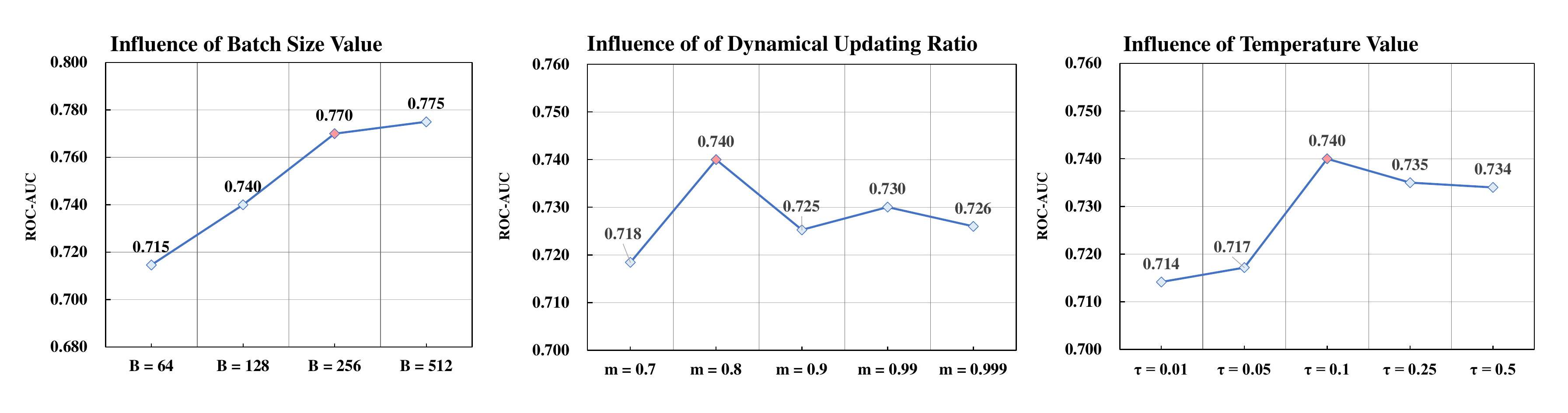}
    \vspace{-5pt}
	\caption{Impact of the value of batch size \begin{math}
        B
    \end{math}, the dynamical updating ratio \begin{math}
        m
    \end{math} and temperature \begin{math}
        \tau
    \end{math} on the result downstream 
 BBBP task. }
	\label{fig:Fig 7.}
\end{figure*}

To assess DIG-Mol’s capabilities, we conducted an experiment to optimize pre-trained hyperparameters. This involved a comparative analysis of fine-tuning models on the BBBP dataset. Details of hyper-parameters under investigation are documented in Table \ref{tab:Tab 6.}. The ultimate results are depicted in Figure \ref{fig:Fig 7.}.

\begin{table}[!ht]
    \centering
    \caption{Details of hyper-parameters under investigation in parameter study.}
    \label{tab:Tab 6.}
    \setlength{\tabcolsep}{2.5mm}
    \begin{tabular}{ccl}
    \toprule[1pt]
    Name & Derivation & Range \\ \hline
    \begin{math}
        B
    \end{math} & Eq.18 & \begin{math}
        \left \{{64, 128, \mathbf{256}, 512} \right \}
    \end{math}  \\ 
    \begin{math}
        m
    \end{math} & Eq.11 & \begin{math}
       \left \{ 0.7, \mathbf{0.8}, 0.9, 0.99, 0.999\right \} 
    \end{math}\\ 
    \begin{math}
        \tau
    \end{math} & Eq.18 & \begin{math}
        \left \{0.01, 0.005, \mathbf{0.1}, 0.25, 0.5 \right \}
    \end{math} \\  \hline
        \end{tabular}
\end{table}

(1) Impact of batch size \begin{math}
        B
\end{math}: Theoretically analysis, a larger batch size should yield improved experimental results. This observation aligns with our experimental findings. However, transitioning from a batch size of 256 to 512 did not yield the expected improvements. The anticipated improvements do not materialize with a larger increase in \begin{math}
        B
\end{math}. Consequently, we fixed \begin{math}
        B
\end{math} at 256 for all classification and regression benchmarks to optimize computational resource efficiency. Due to limitations imposed by the specifications of our experimental equipment, due to equipment limitations, we could not explore further options. Thus, subsequent parameter studies will use a batch size of 128.

(2) Impact of the rate of dynamical updating \begin{math}
    m
\end{math}: Being a critical hyper-parameter in Pseudo-siamese neural networks, \begin{math}
    m
\end{math} influences the update rate of the target network and the quality of molecular representations during pre-training. Our observations reveal that selecting extremely high or low values for \begin{math}
    m
\end{math} is suboptimal. When \begin{math}
    m
\end{math} approaches 1, the online and target networks become too similar, essentially transforming the model into a parameter-sharing siamese network. This erodes the advantage of momentum distillation offered by the pseudo-siamese network, hampers the dual-interaction mechanism, and renders sophisticated design efforts futile. Conversely, excessively small \begin{math}
    m
\end{math} yields undesirable results due to sluggish parameter transmission between the two networks. This hinders the online network’s acquisition of critical information during distillation, limiting comprehensive molecular representation within the finite training epochs. The optimal \begin{math}
    m
\end{math} value falls between 0.8 and 0.9.

(3) Impact of Temperature parameter \begin{math}
    \tau
\end{math}: It is a hyperparameter in the contrastive loss computation function, regulating the model’s ability to distinguish negative samples in contrastive learning. Similar to \begin{math}
    m
\end{math}, excessively high or low values of \begin{math}
    \tau
\end{math} can detrimentally affect the model’s contrastive learning efficacy. In DIG-Mol, \begin{math}
    \tau
\end{math} influences the three types of cross-loss generated by the dual-interaction mechanism, directly impacting contrastive learning quality. When \begin{math}
    \tau
\end{math} is set to an extremely high level, the model treats all negative samples indiscriminately, failing to discern intrinsic differences between samples during the contrastive learning. Conversely, exceedingly low \begin{math}
    \tau
\end{math} values cause the model to overly focus on discriminating intricate negative samples, potentially misclassifying them as positive samples. This misallocation leads to training instability, convergence difficulties, and reduced generalization capabilities. Based on empirical findings, the optimal \begin{math}
    \tau
\end{math} value is approximately 0.1.

\subsection{Ablation study}
\begin{figure}[htb]
    \centering
    \includegraphics[scale=0.16]{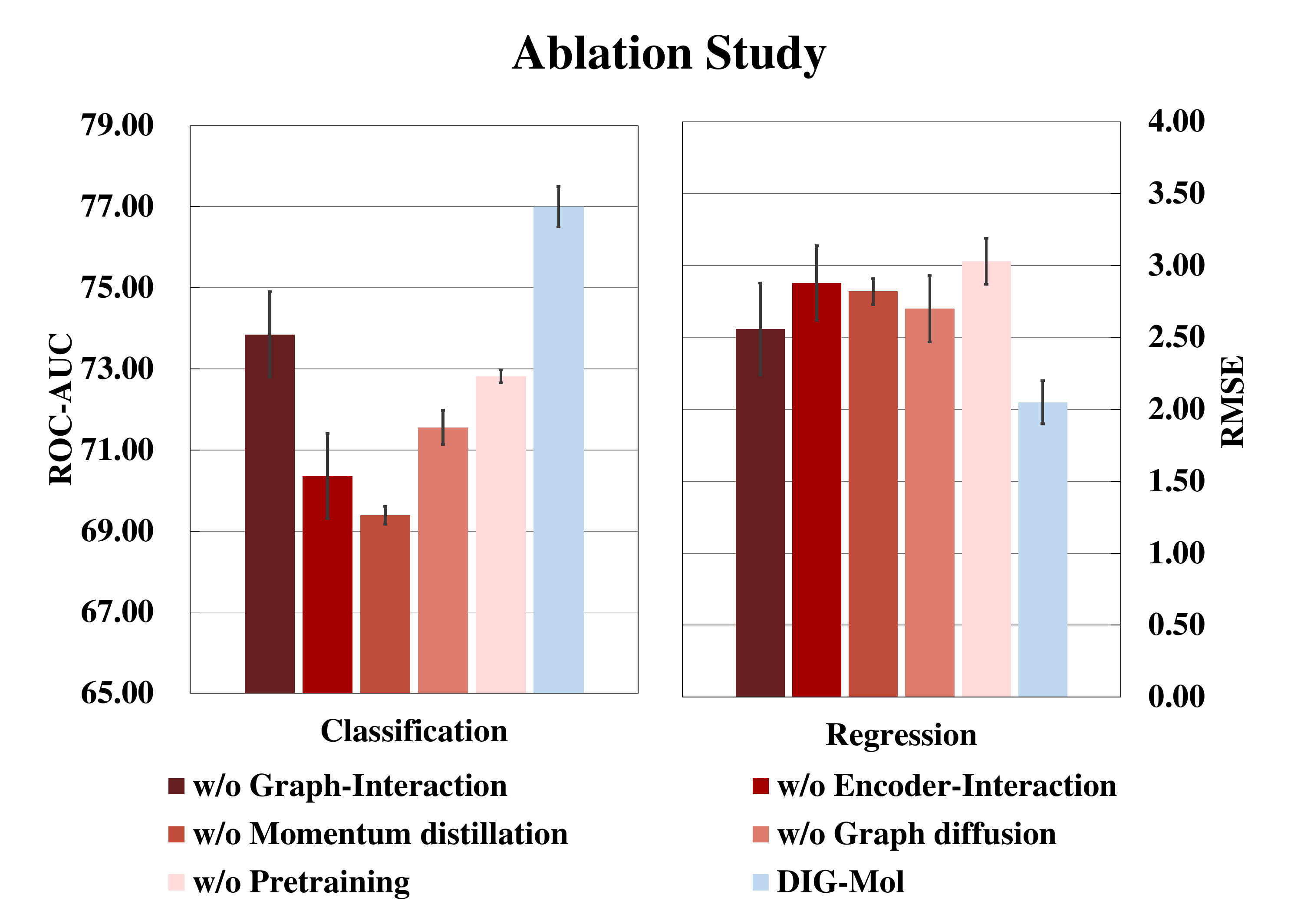}
    \caption{Results obtained from various DIG-Mol variants on both classification and regression tasks.}
    \label{fig:Fig 8}
\end{figure}

To comprehensively assess the impact of integral modules in the DIG-Mol model during pretraining, we conducted an ablation study by comparing DIG-Mol with its variants, each omitting specific modules. The corresponding model variants were labeled as w/o Graph-Interaction, w/o Encoder-Interaction, w/o Momentum distillation, w/o Graph diffusion, and w/o Pretraining. We evaluated these models using BBBP and FreeSolv datasets for representative classification and regression tasks, where the metrics were ROC-AUC and RMSE, respectively. The results are shown in Figure \ref{fig:Fig 8}.

\begin{table}[htb]
    \centering
    \setstretch{1.25}
    \setlength{\tabcolsep}{1.8mm}
    \caption{Results obtained from various DIG-Mol variants on both classification and regression tasks. Mean and standard deviation of test ROC-AUC (\%) on each benchmark are reported.}
    \begin{tabular}{l|cc}
        \toprule[1pt]
        ~ & Classification & Regression \\
        ~ & ROC-AUC & RMSE \\ \hline
        w/o Graph-Interaction & 73.85 ± 1.05 & 2.56 ± 0.32 \\ 
        w/o Encoder-Interaction & 70.36 ± 0.63 & 2.88 ± 0.26 \\ 
        w/o Momentum distillation & 69.39 ± 0.22 & 2.82 ± 0.09 \\ 
        w/o Graph diffusion & 71.56 ± 0.42 & 2.70 ± 0.23 \\
        w/o Pretraining & 72.82 ± 0.16 & 3.03 ± 0.16\\
        DIG-Mol & \textbf{77.00 ± 0.50} & \textbf{2.05 ± 0.15}\\ \hline
    \end{tabular}
    \raggedright
            *The performances of unabridged DIG-Mol are marked as bold
\end{table}

Contrast between w/o Graph-Interaction and w/o Encoder-Interaction: Results showed that w/o Encoder-Interaction exhibit a significantly wider disparity in both classification and regression tasks compared to unabridged DIG-Mol, emphasizing the importance of the unique \begin{math}
    \mathcal{L}_{EI}
\end{math} from Encoder-Interaction. Encoder-Interaction holds greater significance in the optimization of contrastive loss and representation learning, surpassing the \begin{math}
    \mathcal{L}_{GI}
\end{math}, which is commonly employed in regular graph contrastive learning frameworks. Meanwhile, the results generated by w/o Graph-Interaction are comparable to the similar baseline models, providing further evidence to substantiate the indispensable role of Encoder-Interaction and \begin{math}
    \mathcal{L}_{EI}
\end{math} in molecular graph contrastive learning.

Experiment with w/o Momentum distillation: This variant is realized by eliminating Momentum distillation between networks, resulting in parameter-shared networks functioning as siamese networks. However, this alteration yielded unsatisfactory results, suggesting that conventional Siamese networks struggle to perform well in self-supervised learning   with abundant unlabeled data. By creating a supervisory relationship between networks, Momentum distillation facilitates the extraction of critical and comprehensive information from the target network, thereby improving the effectiveness of molecular representation learning.

Examination of w/o Graph diffusion: In this instance, a standard graph neural network was employed as the graph encoder in the w/o Graph diffusion model, yielding results of mediocre quality. This outcome compellingly underscores the critical role of graph diffusion, which significantly enhances the acquisition of molecular graph representations during the learning process.

Exploration of the necessity of pretraining: We trained the w/o Pretraining from scratch on downstream tasks without pretraining weights, resulting in less favorable outcomes. This underscores the crucial role of pretraining in enhancing the model’s capabilities. Aligned with existing literature and empirical findings, the effectiveness of self-supervised pretraining strategies becomes evident when applied to extensive datasets, implicitly acquiring additional molecular information and improving model generalization and transferability.

\end{document}